\PassOptionsToPackage{hyphens}{url}
\documentclass[11pt,a4paper]{article}
\usepackage[a4paper,margin=2.5cm]{geometry}
\usepackage{times}
\usepackage[T1]{fontenc}
\usepackage[utf8]{inputenc}
\usepackage{amsmath,amssymb,booktabs,graphicx,tabularx,multirow}
\usepackage{microtype}
\usepackage{xcolor}
\usepackage[font=small]{caption}
\usepackage{natbib}
\usepackage{url}
\usepackage[colorlinks=true,linkcolor=blue!50!black,citecolor=blue!50!black,urlcolor=blue!50!black]{hyperref}
\graphicspath{{figures/}}

\newcommand{\R}{\mathbb{R}}
\newcommand{\pred}{\Pi_{\mathrm{pred}}}
\newcommand{\inp}{\Pi_{\mathrm{input}}}
\newcommand{\beh}{\Pi_{\mathrm{behav}}}
\newcommand{\rsf}{\operatorname{RSF}}
\newcommand{\srank}{\operatorname{srank}}
\newcommand{\tr}{\operatorname{tr}}

\title{The Geometry of Learning: Read-Write Asymmetry Across Training}
\author{Valeria Ruscio, Keiran Thompson}
\date{Intuition Machines Inc}

\begin{document}
\maketitle

\begin{abstract}
Pretraining and alignment update the same transformer weights, and we ask in which directions each stage changes them. The residual stream gives every matrix a shared coordinate system, in which a weight change can be compared with the principal subspace of the activations the matrix reads and with the prediction subspace of the unembedding. In this frame, alignment updates to the query and key matrices are enriched in the principal input subspace, while its updates to the output matrices are close to random relative to the prediction subspace, and the same signature appears in late pretraining and under every objective we apply to one checkpoint, so it is a property of the training stage rather than of the objective. The gradient of a linear map is a sum of outer products $\delta_t a_t^\top$, and the residual stream enters this product through the activation $a_t$ for matrices that read from it and through the upstream gradient $\delta_t$ for matrices that write into it. Read gradients therefore inherit the spectrum of the activations, which is the same under every objective, while write gradients inherit the geometry of the upstream gradients, which point into the prediction subspace only early in pretraining, when the output matrices acquire their prediction geometry. The raw read gradient follows the activation spectrum, the optimiser attenuates the concentration it inherits, and cancellation between tokens sets the scale of the update but not its direction. A write direction derived from the unembedding still changes predictions after alignment, and adaptation restricted to any single group of matrices remains effective, so the signature describes where training places its updates, not which matrices can carry them.
\end{abstract}

\section{Introduction}\label{sec:intro}
Large language models are trained in stages, with cross-entropy pretraining on large text corpora followed by alignment through supervised fine-tuning (SFT) and preference optimisation such as DPO and ORPO \citep{rafailov2023direct,hong2024orpo}. The two stages act on the same weights, but they are usually compared through their effect on behaviour, and when their effects on the weights are compared at all, the comparison is about how much a matrix changes, through the norm of the update or its rank, since fine-tuning updates are known to live in a low intrinsic dimension \citep{aghajanyan2021intrinsic,hu2021lora} and contextual representations are known to be anisotropic \citep{ethayarajh2019contextual}. A change of low rank, however, can point in any direction, and what it does to the model depends on which directions it occupies, because the modified matrix acts on the rest of the network through them. The question we ask is therefore not how much pretraining and alignment change each matrix, but in which directions, and why. It matters for interpretability, because the direction of a change says which part of the computation a training stage modified, in practice, because model editing, adapter placement and model merging all rely on assumptions about where knowledge and behaviour are stored \citep{rome,hu2021lora,ilharco2022editing}, and for the debate about whether alignment mostly changes how a model uses what it learned in pretraining \citep{zhou2023lima,lin2024unlocking,qi2024safetyalignmentjusttokens}, which has so far been examined through behaviour rather than through the weights.

\begin{figure}[t]
\centering
\includegraphics[width=0.9\textwidth]{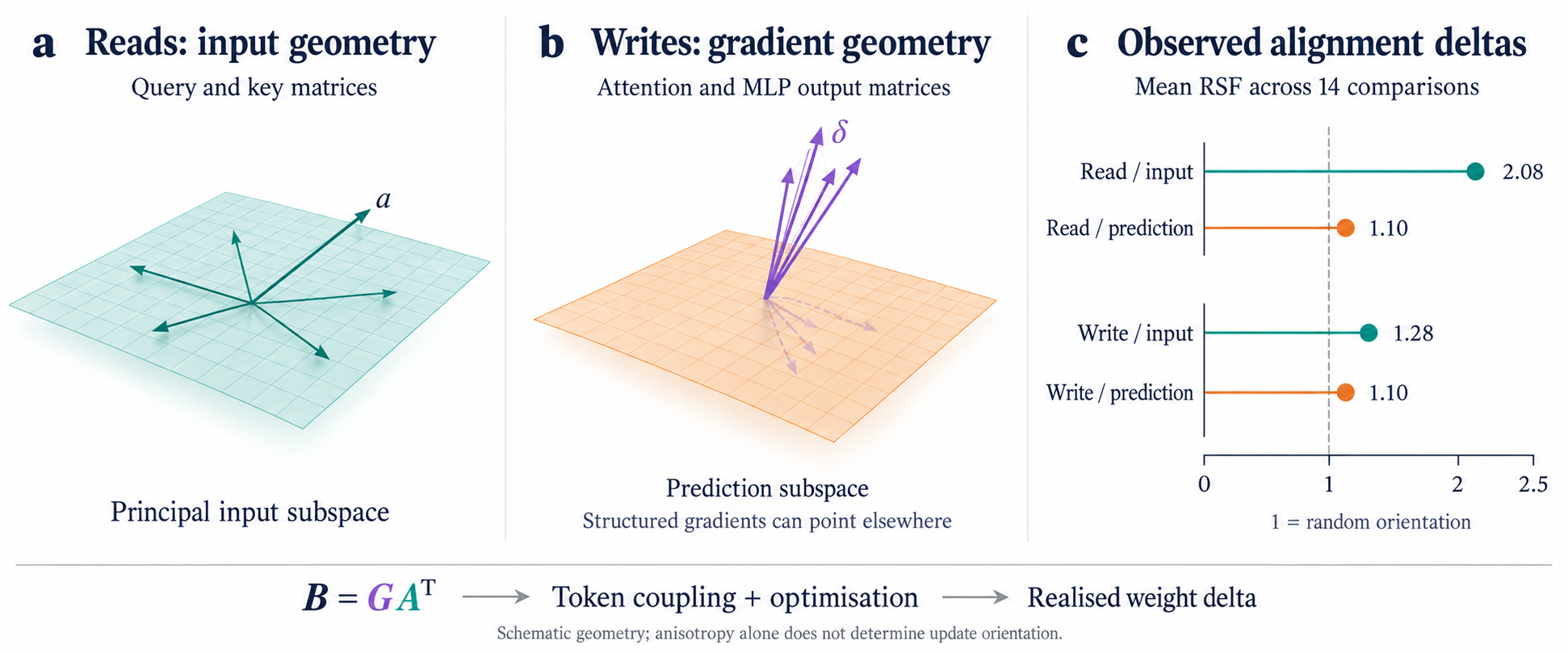}
\caption{Two sides of gradient accumulation. (a) Read geometry is referenced to principal input activations (teal). (b) Structured upstream gradients (purple) need not align with the prediction subspace (orange); both sketches are illustrative. (c) Mean alignment-delta RSF across 14 checkpoint comparisons, reads relative to inputs and writes relative to prediction directions, where one denotes random orientation. Both pathways obey $B=GA^\top$, but token coupling and optimisation determine the realised weight delta.}
\label{fig:pathways}
\end{figure}

Comparing directions requires a coordinate system that matrices, models and training stages share, and the residual stream provides one, because every layer reads its input from it and writes its output back into it \citep{circuits}. The query and key matrices $W_Q,W_K$ read from the stream to compute attention scores, and we call them the \emph{read pathway}, while the attention and MLP output matrices $W_O,W_2$ write into it, and we call them the \emph{write pathway}. Each pathway has a natural reference subspace, the principal subspace of the activations it reads for the read pathway, and the subspace through which the unembedding maps the residual stream most strongly onto the logits for the write pathway \citep{geva2021transformer,belrose2023eliciting}, and the share of a weight change that falls in a reference subspace, compared with the share expected under a random orientation, measures where the change points independently of its size.

The same frame also says why the two pathways might be changed in different directions by the same training stage. The gradient of any linear map is a sum of outer products $\delta_t a_t^\top$ of an upstream gradient and an input activation, and the residual stream enters these products on opposite sides for the two pathways, through the activations $a_t$ for reads and through the upstream gradients $\delta_t$ for writes. The geometry of a read update is therefore tied to the statistics of the activations, which are the same for any objective applied to the same model and data, while the geometry of a write update is tied to the statistics of the upstream gradients, which depend on the objective and on the stage of training. Whether the realised weight changes follow these statistics, and how closely, is a question of measurement, and it is the question of this paper. We ask where alignment deltas point relative to the two reference subspaces, whether the answer is specific to alignment or shared by any late update, how much of the raw gradient's enrichment the realised update keeps, whether the write updates are structured and where they point, when during pretraining the write pathway acquires its relation to the prediction subspace, and what the resulting geometry leaves open to editing and adaptation.

\section{Related work}\label{sec:related}
\paragraph{Residual-stream and prediction geometry}
\citet{circuits} describe the residual stream as a shared communication channel between attention and MLP sublayers, which is the view our pathways build on, the key--value view of MLPs connects their output vectors to predictions in vocabulary space \citep{geva2021transformer}, and the tuned lens studies how intermediate representations encode predictions \citep{belrose2023eliciting}. Representation anisotropy is well documented \citep{ethayarajh2019contextual}, but spectral concentration alone does not say which reference subspace a representation or an update occupies, and RSF measures exactly this relative orientation, so that a delta can be low rank yet nearly orthogonal to $\pred$, or high rank while keeping above-random mass in $\inp$.

\paragraph{What alignment changes}
The view that alignment mostly changes how a model uses knowledge acquired in pretraining rests on the observations that a small instruction set suffices to align a strong base model \citep{zhou2023lima} and that aligned and base models differ mainly on stylistic tokens \citep{lin2024unlocking}; its scope has been revisited as post-training budgets grow \citep{raghavendra2024revisiting}, and safety alignment in particular acts mainly on the first few output tokens \citep{qi2024safetyalignmentjusttokens}. Our results give these observations a weight-space counterpart: what distinguishes alignment in weight space is when it happens rather than what it optimises.

\paragraph{Model editing and adaptation}
ROME and MEMIT modify factual associations through targeted changes to MLP weights \citep{rome,meng2022mass}, LEACE removes linearly decodable concepts \citep{belrose2023leace}, and \citet{arditi2024refusal} ablate a direction by weight orthogonalisation, $W\leftarrow W-uu^\top W$, the operation we use with partial attenuation and unembedding-derived directions, while LoRA constrains the parameterisation of the update \citep{hu2021lora} and task arithmetic and TIES-Merging ask how independently learned deltas compose \citep{ilharco2022editing,yadav2023ties}.

\section{Measuring read and write geometry}\label{sec:methods}
\subsection{Pathways and reference subspaces}
Let $h^{(\ell)}\in\R^d$ denote the residual stream at layer $\ell$, with the column-vector convention $y=Wa$. The read-pathway matrices $W_Q,W_K$, which we also call direct reads, have residual-stream coordinates on their input side, while the write-pathway matrices $W_O\in\R^{d\times d_v}$ and $W_2\in\R^{d\times d_{\mathrm{ff}}}$ have them on their output side. The value and MLP input matrices $W_V,W_1$ also read the normalised residual stream, but reach it again only through a subsequent output map, and we call them \emph{internal reads}; for gated MLPs this tier includes the gate and the up projections.

The unembedding $W_U\in\R^{V\times d}$ maps final representations to vocabulary logits, and if $V_k$ contains its leading $k$ right-singular vectors, the prediction projector is $\pred=V_kV_k^\top$, which selects the directions along which the unembedding has the largest gain. The input projector $\inp$ is estimated at each layer from the principal directions of the layer-input activations, by default using base-model activations on WikiText-2, with other corpora and aligned-model activations tested in Section~\ref{sec:robust}. A third projector, $\beh$, is formed from the principal components of final-layer activation differences between the aligned and the base models, and serves as a reference tied to the behavioural change itself.

\subsection{Relative subspace fraction}
For a nonzero matrix $M$ and a rank-$k$ orthogonal projector $\Pi\in\R^{d\times d}$, we define
\begin{equation}
 f(M,\Pi)=
 \begin{cases}
 \|\Pi M\|_F^2/\|M\|_F^2,&\text{writes},\\
 \|M\Pi\|_F^2/\|M\|_F^2,&\text{reads},
 \end{cases}
 \label{eq:rsf}
\end{equation}
and $\rsf(M,\Pi)=f(M,\Pi)/(k/d)$, where the projector always acts on the residual-stream side of the matrix, on the columns of a write and on the rows of a direct or internal read. The raw fraction $f$ lies in $[0,1]$, and RSF has expectation one under a rotationally uniform orientation independent of $\Pi$, so an RSF of $2$ means twice the random-orientation share. We use $k=50$ unless stated otherwise, and since RSF is unchanged when $M$ is multiplied by a nonzero scalar, it measures orientation independently of magnitude; an RSF of $2$ at $d=4096$ is still only $2.4$ percent of the delta energy, so we describe deltas as \emph{enriched} in a subspace rather than as lying in it. To compare two subspaces with orthonormal bases $U,V\in\R^{d\times k}$ we use the overlap
\begin{equation}
 \mathcal O(U,V)=\frac{\|U^\top V\|_F^2}{k}
 =\frac{\tr(UU^\top VV^\top)}{k},
 \label{eq:overlap}
\end{equation}
the mean squared cosine of the principal angles, equal to zero for orthogonal and one for identical subspaces, with expectation $k/d$ for independent random subspaces.

\subsection{Models and checkpoint comparisons}\label{sec:cohorts}
The static protocol applies Equation~\ref{eq:rsf} to the alignment delta $\Delta W=W_{\mathrm{aligned}}-W_{\mathrm{base}}$, and for two-stage pipelines we also isolate $W_{\mathrm{DPO}}-W_{\mathrm{SFT}}$, since a base-to-DPO delta can otherwise mainly describe the preceding SFT stage. The static cohort contains seven base models from four open families, Llama~3, including the Tulu~3 pipeline, OLMo~2, Qwen2.5 and Mistral, each paired with publicly released SFT, DPO or ORPO checkpoints (Appendix~\ref{app:cohorts}). Within each checkpoint comparison we average over layers and over the matrices of a tier, and Table~\ref{tab:static} gives equal weight to each of the 14 base-to-aligned comparisons. The pretraining protocol instead measures the full weight $W^{(t)}$ against the projector $\pred^{(t)}$ of the same checkpoint, checked with interval deltas between consecutive checkpoints, which separate the movement of the weights from that of the reference subspace. Upstream gradients are measured in Pythia-410M \citep{biderman2023pythia} and OLMo-2-1B, and attention-score gradients in a third cohort of Llama, OLMo and Qwen checkpoints and five public pretraining ladders (Appendix~\ref{app:selection-details}). Two controlled protocols on Pythia-410M complement the static cohort, an interval delta between two late pretraining checkpoints, and three short training branches, continued cross-entropy, SFT and DPO, grown from one warmed checkpoint with the same optimiser, batch and step count, along which raw gradients, optimiser steps and cumulative deltas are measured (Appendix~\ref{app:controlled}). Code will be released.

\section{Anisotropic gradient accumulation}\label{sec:mechanism}
\subsection{Two sides of the gradient}
For a linear map $y_t=Wa_t$, the contribution of token $t$ to the raw parameter gradient is the outer product $\delta_t a_t^\top$, where $\delta_t=\partial\mathcal L/\partial y_t$ is the upstream gradient, and stacking token inputs and upstream gradients as columns gives
\begin{equation}
 \begin{gathered}
 A=[a_1,\ldots,a_N],\quad G=[\delta_1,\ldots,\delta_N],\\
 B=\nabla_W\mathcal L=GA^\top,
 \end{gathered}
 \label{eq:gradient}
\end{equation}
with the weights of the loss reduction absorbed into the factors. The column and row spaces of this product satisfy
\begin{equation}
 \operatorname{col}(B)\subseteq\operatorname{col}(G),\qquad
 \operatorname{row}(B)\subseteq\operatorname{row}(A^\top).
 \label{eq:containment}
\end{equation}
This identity locates the residual-stream coordinates of each gradient: for a read-pathway matrix they appear in the input factor $A$, and for a write-pathway matrix in the upstream-gradient factor $G$, so the residual-side directions of a read gradient are constrained by the activations, and those of a write gradient by the upstream gradients.

\subsection{Candidate directions and realised updates}
The containments identify the subspace in which a gradient can lie, but not how its energy is distributed inside it, which also depends on how the two factors are coupled across tokens. Writing $A=U_AS_AV_A^\top$ and $G=U_GS_GV_G^\top$ gives
\begin{equation}
 B=U_GS_G\underbrace{(V_G^\top V_A)}_{\text{token coupling}}S_AU_A^\top,
 \label{eq:svd}
\end{equation}
where the singular spectra of $A$ and $G$ provide candidate directions of concentration, and $V_G^\top V_A$ determines how token contributions combine, so that a high-variance activation direction can receive little update energy when its gradient-weighted contributions cancel. Leakage outside a reference subspace is nevertheless controlled by the factors, since for a read-side projector $P$ and a write-side projector $Q$
\begin{equation}
 \begin{aligned}
 \|B(I-P)\|_F&\leq\|G\|_2\,\|(I-P)A\|_F,\\
 \|(I-Q)B\|_F&\leq\|(I-Q)G\|_F\,\|A\|_2,
 \end{aligned}
 \label{eq:bounds}
\end{equation}
so small leakage in the relevant factor bounds the leakage of the gradient, while a guarantee on normalised concentration would also require control of $\|B\|_F$, which cancellation can make small. A checkpoint difference adds one more layer between the factors and the realised update, since plain gradient descent accumulates $-\eta B$ along a moving trajectory, while adaptive optimisers rescale coordinates, momentum combines gradients across time, and weight decay adds a component proportional to the current weights. The rest of the paper tests one statement: the anisotropy of the relevant factor supplies the candidate directions of an update, and token coupling and optimisation decide how much of each candidate is realised.

\subsection{What the two factors look like}
\paragraph{Read pathway}
For $W_Q$ and $W_K$ the input factor is the normalised residual-stream activation, whose spiked spectrum gives read gradients a concentrated reference under every objective, while the upstream factor reaches the weights only through the attention scores. For query position $i$ with attention weights $m_i=\operatorname{softmax}(s_i)$,
\begin{equation}
 \frac{\partial\mathcal L}{\partial s_i}=J_i\,\frac{\partial\mathcal L}{\partial m_i},\qquad J_i=\operatorname{diag}(m_i)-m_im_i^\top,
 \label{eq:softmax}
\end{equation}
and since $J_i\mathbf 1=0$ and $\|J_i\|_2\leq 1/2$, a change that raises all scores equally receives no gradient, so each token contributes only a contrast among candidate keys, and contrasts from different tokens need not agree. Internal reads share the input side, but their gradient passes through additional downstream maps, which is consistent with their weaker enrichment.

\paragraph{Write pathway}
For $W_O$ and $W_2$ the relevant statistics are those of the upstream gradient in residual-stream coordinates, where two distinct properties matter, whether the gradient spectrum is concentrated and whether its concentrated directions align with $\pred$, since strong anisotropy can occur outside the prediction subspace. For cross-entropy, the unembedding maps the logit error directly into the gradient of its input, since for target $y$, logits $z$ and probabilities $p$ the logit gradient is $p-e_y$, with $e_y$ the one-hot vector of the target, and at the representation $h_U$ that enters the unembedding
\begin{equation}
 \nabla_{h_U}\mathcal L=W_U^\top(p-e_y).
 \label{eq:ce}
\end{equation}
The one-hot target ties each example to a single vocabulary item, but the covariance of this signal across examples, and its orientation in the singular basis of the unembedding, depend on both targets and predictions, so the objective fixes the form of the write-side factor without fixing its orientation, and since the Jacobians between the unembedding and earlier write outputs transform it further, prediction alignment must be measured at the write outputs.

\subsection{Implications for the realised updates}
The point of this analysis is that the orientation of an update follows from quantities that are measurable on their own, the spectrum of the activations for the read pathway and the spectrum and orientation of the upstream gradients for the write pathway, so that the realised update can be compared with them to see how much of the candidate geometry survives coupling and optimisation. Four consequences follow. Since the input factor does not depend on the loss, concentrated input activations accompany the read signature in every setting and under every objective. Read gradients cancel heavily across tokens, which bounds the norm of the summed gradient but not its orientation, so any departure of the realised update from the activation spectrum comes from the coupling of the factors or from the optimiser, and the controlled branches separate the two. For the write pathway, the mechanism requires that the gradients alignment sends to the write outputs are structured but oriented away from $\pred$, and that during pretraining the write weights concentrate in the prediction subspace at the time when the upstream gradients point into it. Section~\ref{sec:static} establishes the signature, and Sections~\ref{sec:reads} and~\ref{sec:dynamics} verify these four consequences.

\section{Where do alignment deltas point?}\label{sec:static}
\subsection{A pathway--subspace dissociation}
In the matched alignment deltas of Table~\ref{tab:static}, read-pathway deltas have mean RSF $2.08$ in $\inp$ but only $1.10$ in $\pred$, and read/$\inp$ exceeds read/$\pred$ in every one of the 14 comparisons, with read/$\inp$ values from $1.22$ to $3.14$. Write-pathway deltas have mean RSF $1.10$ in $\pred$, with values from $1.03$ to $1.21$, and they are more enriched in $\inp$ ($1.28$) than in the prediction subspace, so alignment changes the write pathway too, in a direction its own reference subspace does not capture. In a mixed-effects model of log RSF with a random intercept per base model, the contrast between the pairing the mechanism assigns and the other references in the read pathway is $\beta=0.98$ ($z=5.89$, $p=4\times10^{-9}$), and a one-sided exact Wilcoxon signed-rank test with one contrast per base model gives $p=0.008$, the smallest value attainable with seven units, since the contrast is positive in all of them. The corresponding write contrast is negative ($\beta=-0.18$, $z=-6.81$), so writes are less enriched in $\pred$ than in the other references taken together. Since the seven base models come from four families and contribute dependent layer observations, the effect sizes and their consistency carry more weight than the layer-level $p$-values (Appendix~\ref{app:stats}, Table~\ref{tab:static-full}).

\begin{table}[t]
\centering
\small
\begin{tabular}{lccc}
\toprule
Tier & $\pred$ & $\inp$ & $\beh$\\
\midrule
Writes: $W_O,W_2$ & $1.10\pm0.05$ & $1.28\pm0.08$ & $1.09\pm0.05$\\
Reads: $W_Q,W_K$ & $1.10\pm0.11$ & $\mathbf{2.08\pm0.61}$ & $1.11\pm0.12$\\
Internal: $W_V,W_1$ & $1.06\pm0.06$ & $1.36\pm0.23$ & ---\\
Random & $1.00$ & $1.00$ & $1.00$\\
\bottomrule
\end{tabular}
\caption{RSF of the static alignment deltas, mean and sample standard deviation over the 14 base-to-aligned comparisons (Table~\ref{tab:static-full}); internal reads use the 12 comparisons with an internal-read measurement (Table~\ref{tab:internal}), averaging $W_V$ and the gate and up projections.}
\label{tab:static}
\end{table}

The internal reads complete a three-tier ordering, since their deltas are enriched in $\inp$ with mean RSF $1.36$, below direct reads and above writes, while staying close to random in $\pred$, and within the MLP the gate projection, which decides which features are active, exceeds the up projection in $\inp$ in 10 of these 12 comparisons, the two exceptions being ORPO checkpoints. Isolating the second stage of a two-stage pipeline can reveal structure that a base-referenced delta hides. For OLMo-2-7B, the base-to-SFT and base-to-DPO read/$\inp$ ratios are $1.71$ and $1.70$, while the isolated DPO-minus-SFT delta gives $2.67$, so a strongly oriented DPO update sits inside a base-referenced summary that barely changes, whereas for Tulu the three values are $2.17$, $2.18$ and $2.08$.

\subsection{Orientation is not size, rank or inherited structure}\label{sec:robust}
The read signature survives the choice of rank, $k\in\{32,50,64\}$, calibration on WikiText-2, C4, OASST1 and Alpaca, where it is largest for instruction data, and projectors built from aligned instead of base activations, which change the read ratio by less than $0.07$ (Tables~\ref{tab:rank}--\ref{tab:source}).
Nor is it inherited from the weights, since OLMo-2 base-to-SFT deltas are enriched in the leading singular subspaces of their own base weights in both pathways, with ratios $8.63$ for reads and $4.22$ for writes, yet keep a read ratio of $2.01$ against write ratios of $0.89$--$0.98$ in the activation reference, so the signature does not restate the fact that updates are low-rank and weights are structured.

\subsection{The same signature in late pretraining and under every objective}\label{sec:notspecific}
The signature of Table~\ref{tab:static} is measured on alignment deltas, but the mechanism of Section~\ref{sec:mechanism} implies that it is not specific to them, since the input factor is the same under every objective, and we test this on Pythia-410M in two ways. The first is a late pretraining interval, from step 128,000 to the final step 143,000, with input projectors built at the start of the interval. The interval delta has read/$\inp$ RSF $3.30$, with a descriptive layer-bootstrap interval of $[2.30, 4.38]$, all 24 layers lie above one, the raw projected fraction is $0.16$, and the enrichment survives changes of rank and centering (Table~\ref{tab:late}). The second is a controlled comparison of objectives. From the final checkpoint we take 64 SFT steps on preference data to obtain a warmed anchor, and from the anchors of three seeds we grow three branches of 128 steps each with the same optimiser, batch and step count, continued cross-entropy on WikiText-2, SFT on the chosen responses, and DPO (Appendix~\ref{app:controlled}). All three branches are enriched in the input subspace of the read pathway, read/$\pred$ is $1.41$ to $1.46$ in the three, and the write pathway's enrichment in $\pred$ is $1.42$ in all three, identical to two decimals, while write/$\inp$ is larger under continued cross-entropy ($1.82$) than under SFT and DPO ($1.56$), so the objective moves the write pathway within the input subspace and not within the prediction subspace (Table~\ref{tab:controlled}). The read enrichment is largest under continued cross-entropy, which is confounded with the calibration domain, since the branch and the projectors both use WikiText-2, and the DPO branch does not reach consistent behavioural alignment, with a relative preference accuracy over its anchor of 46 to 55 percent (Table~\ref{tab:controlled-eval}), so the comparison shows that each objective produces the signature, not that the differences between objectives are objective effects. Together the two protocols show that the signature is a property of training after the write geometry has formed, not of alignment.

\begin{table}[t]
\centering
\small
\begin{tabular}{lcccc}
\toprule
Branch & Rd/$\inp$ & Rd/$\pred$ & Wr/$\inp$ & Wr/$\pred$\\
\midrule
Text CE & $6.91\pm0.05$ & $1.41\pm0.01$ & $1.82\pm0.02$ & $1.42\pm0.02$\\
SFT & $3.52\pm0.07$ & $1.46\pm0.01$ & $1.56\pm0.01$ & $1.42\pm0.00$\\
DPO & $3.74\pm0.01$ & $1.44\pm0.01$ & $1.56\pm0.01$ & $1.42\pm0.01$\\
\bottomrule
\end{tabular}
\caption{RSF of the branch deltas in the controlled comparison on Pythia-410M (Rd: reads, Wr: writes), rank 50, mean $\pm$ sample standard deviation over three seeds, each averaging the matrices of a tier within layers and then layers equally.}
\label{tab:controlled}
\end{table}

\section{The read pathway: inputs, cancellation and the optimiser}\label{sec:reads}
Section~\ref{sec:mechanism} attributes the read signature to the concentration of the input factor, and if the summed gradient followed that factor, the principal input directions would receive update energy in proportion to their variance. The realised updates receive far less, and this section locates the difference.

\subsection{The input side is concentrated in every setting}
At the post-normalisation attention inputs of OLMo-2-1B, the top 50 activation directions carry $0.30$--$0.42$ of the energy, against an isotropic baseline of $0.024$, in every setting we measured, generic-text cross-entropy, SFT, DPO and instruction-tuned, and at the same site in Pythia-410M they carry $0.36$--$0.72$ across layers, with a mean of $0.44$. On a given input distribution, the principal input subspaces of base and aligned models largely coincide, with mean principal-angle cosines of $0.889$ in Llama-3.2-3B and $0.886$ in Mistral-7B, against about $0.1$ for random subspaces, so both present the same anisotropic reference.

\subsection{The raw gradient follows the activations; the optimiser does not}\label{sec:optimiser}
For an activation eigenpair $(q_i,\lambda_i)$ of the input covariance $\Sigma_a$ we define the allocation ratio
\begin{equation}
 c_i=\frac{\|\Delta Wq_i\|_2^2/\|\Delta W\|_F^2}{\lambda_i/\tr(\Sigma_a)},
 \label{eq:allocation}
\end{equation}
which equals one when $q_i$ receives the same share of update energy as of activation variance. In six aligned checkpoints of the static cohort, measured at one layer, the mean of $c_i$ over the 50 leading directions lies between $0.053$ and $0.156$, while these directions hold $0.43$--$0.61$ of the activation trace, so each dominant direction receives 6 to 19 times less of the delta energy than its share of activation variance, and in aggregate 12 to 29 times less, although the delta still ends above chance in the subspace they span (Table~\ref{tab:allocation}). The controlled branches show where this shortfall arises. At four steps and three layers we measure, for $W_Q$ and $W_K$, the raw gradient, the step the optimiser took, the cumulative delta since the anchor, and the learning-rate-scaled sum of the clipped gradients, which is what plain gradient descent would have accumulated (Table~\ref{tab:diagnostics}). The raw gradient is strongly enriched, with read/$\inp$ RSF $10.0$ under continued cross-entropy and $7.6$ under SFT and DPO, and under cross-entropy its energy in the top-50 input subspace equals the share of activation variance those directions hold, with a ratio of $1.01$ over the probed layers, so the raw gradient does follow the activations. Pairing each upstream gradient with the activation of a permuted token leaves the RSF unchanged, at $9.8$, $7.9$ and $7.9$, so the orientation of the summed gradient is set by the activation factor and not by the coupling of the two factors. The optimiser reduces this enrichment: the Adam step has RSF $5.9$, $3.1$ and $3.1$ and the cumulative delta $6.7$, $3.0$ and $3.1$, while the plain sum of clipped gradients keeps $10.8$, $7.4$ and $7.6$, so coordinatewise normalisation roughly halves the concentration the raw gradient inherits, and the step's energy in the leading directions falls to $0.64$ and $0.31$ of their variance share. The larger shortfall of the static cohort is the same effect accumulated over much longer training.

\begin{table}[t]
\centering
\small
\begin{tabular}{lccccccc}
\toprule
 & \multicolumn{5}{c}{Read/$\inp$ RSF} & \multicolumn{2}{c}{Coherence}\\
\cmidrule(lr){2-6}\cmidrule(lr){7-8}
Branch & Raw & Shuffled & Adam & Delta & $\Sigma$ clipped & In & Out\\
\midrule
Text CE & 10.0 & 9.8 & 5.9 & 6.7 & 10.8 & .066 & .065\\
SFT & 7.6 & 7.9 & 3.1 & 3.0 & 7.4 & .084 & .085\\
DPO & 7.6 & 7.9 & 3.1 & 3.1 & 7.6 & .060 & .061\\
\bottomrule
\end{tabular}
\caption{Read-side gradient and optimiser diagnostics in the controlled branches, averaged over seeds, probe steps, layers and $W_Q,W_K$: the raw gradient, the same factors with permuted token pairing, the optimiser's step, the cumulative delta, and the learning-rate-scaled sum of clipped gradients; coherence is the norm of the summed token contributions over the sum of their norms, inside and outside $\inp$.}
\label{tab:diagnostics}
\end{table}

\subsection{Token contributions cancel, but not selectively}
Cancellation between tokens is strong: the coherence of the read gradient, the norm of the summed token contributions divided by the sum of their norms, is between $0.06$ and $0.09$ in every branch, so more than 90 percent of the per-token norm cancels, and it is the same inside and outside the principal input subspace, with the inside value larger in only 29 to 37 of the 72 probes (Table~\ref{tab:diagnostics}). 
The same disagreement appears among attention decisions in larger models: recording the sign of the score gradient of each query's winning key, 94 to 97 percent of the signs cancel in seven Llama, OLMo and Qwen checkpoints and four corpora, with a residual above the permutation floor confined to individual heads (Table~\ref{tab:selection-static}, Appendix~\ref{app:selection-details}). Cancellation therefore sets the norm of the summed gradient, as the bounds of Equation~\ref{eq:bounds} allow, but not its orientation, and the departure of the realised update from the activation spectrum comes from the optimiser rather than from the coupling of tokens.

\subsection{The read signature under a contrastive objective}\label{sec:objectives}
Four contrastive fine-tuning pairs trained with InfoNCE show that the signature does not depend on cross-entropy either: two encoder backbones with their sentence-transformer variants, MiniLM-L6 and MPNet-base, for which the tied input embeddings take the place of $W_U$, and two decoders, GTE-Qwen2-1.5B and E5-Mistral-7B. Read/$\inp$ ratios exceed one in all four pairs, between $1.28$ and $1.92$ (Table~\ref{tab:contrastive}), and on the matched Mistral-7B base, SFT and InfoNCE give read ratios of $1.77$ and $1.28$, with SFT higher in $W_Q$ and in $W_K$ (Table~\ref{tab:mistral-objective}; paired Wilcoxon $p<10^{-3}$ across 32 layers). The write side of these pairs is the first case in which the pattern is not uniform: three have write/$\pred$ ratios of $1.07$ to $1.10$, in the range of the alignment deltas, while E5-Mistral-7B reaches $1.71$, well above the largest alignment value of $1.21$, so an objective applied after pretraining can produce substantial write-side prediction enrichment.

\section{What shapes the write pathway, and when?}\label{sec:dynamics}
\subsection{Structured gradients that point elsewhere}
The weak prediction enrichment of the write pathway could be read as evidence that alignment sends little structured signal to the write matrices; the mechanism of Section~\ref{sec:mechanism} requires the opposite. We collect $\delta_t=\partial\mathcal L/\partial y_t$ at the outputs of $W_O$ and $W_2$ with backward hooks, keeping the individual token-gradient vectors, with 15,000--123,000 token gradients per cell in OLMo-2-1B (Appendix~\ref{app:gradient}). For an estimated gradient covariance $C_g$ with eigenvalues $\lambda_1\geq\cdots\geq\lambda_d$ and top-$k$ eigenvectors $U_g$, we report
\begin{equation}
 m_k=\frac{\sum_{i=1}^k\lambda_i}{\tr(C_g)},\qquad
 o_k=\frac{\tr(U_gU_g^\top\pred)}{k},
 \label{eq:gradientstats}
\end{equation}
where the first quantity measures concentration and the second the orientation of the leading directions relative to $\pred$, both with reference value $k/d$.

\begin{table}[t]
\centering
\small
\begin{tabular}{llccc}
\toprule
Model & Setting & $m_k$ & $o_k$ & $k/d$\\
\midrule
OLMo-2-1B & Text CE & $0.41$--$0.47$ & $0.031$--$0.036$ & $0.024$\\
 & SFT & $0.55$--$0.59$ & $0.026$--$0.029$ & $0.024$\\
 & DPO & $0.43$--$0.52$ & $0.029$--$0.035$ & $0.024$\\
Pythia-410M & Text CE & $0.57$ & $0.076$ & $0.049$\\
 & SFT & $0.74$ & $0.069$ & $0.049$\\
 & DPO & $0.65$ & $0.071$ & $0.049$\\
\bottomrule
\end{tabular}
\caption{Upstream-gradient statistics at the write outputs (Equation~\ref{eq:gradientstats}), top-50 covariance mass $m_k$ and overlap $o_k$ with $\pred$, with reference value $k/d$. OLMo-2-1B rows are ranges across checkpoints and layers; Pythia-410M rows are means over seeds, probe steps and layers in the controlled branches.}
\label{tab:grads}
\end{table}

The gradients are strongly anisotropic in all three OLMo-2-1B settings of Table~\ref{tab:grads}, with SFT the most concentrated, holding 55 to 59 percent of its covariance in 50 of 2048 directions, and yet the dominant eigenspaces of the alignment gradients overlap $\pred$ at $0.026$--$0.035$, barely above the baseline of $0.024$. Gradient concentration and prediction alignment are different properties, and alignment has the first without the second, so the write pathway receives a structured signal that lies outside the subspace through which the unembedding reads it. The controlled branches reproduce this on Pythia-410M, where the write-output gradients hold $0.57$ to $0.74$ of their covariance in 50 of 1024 directions while overlapping $\pred$ at $0.069$--$0.076$ against a baseline of $0.049$, the same in all three branches to within a hundredth, and the resulting write deltas have the same $\pred$ enrichment of $1.42$ under all three objectives (Table~\ref{tab:controlled}). The same table raises a further question, because the gradients of converged or continued cross-entropy overlap $\pred$ just as weakly. If the pretraining objective at convergence does not point the write gradients into the prediction subspace, the objective alone cannot explain the prediction geometry that trained write matrices do have, and the mechanism places its formation at a stage of training rather than in the objective.

\subsection{The prediction geometry forms early}
\begin{figure}[t]
\centering
\includegraphics[width=0.55\textwidth]{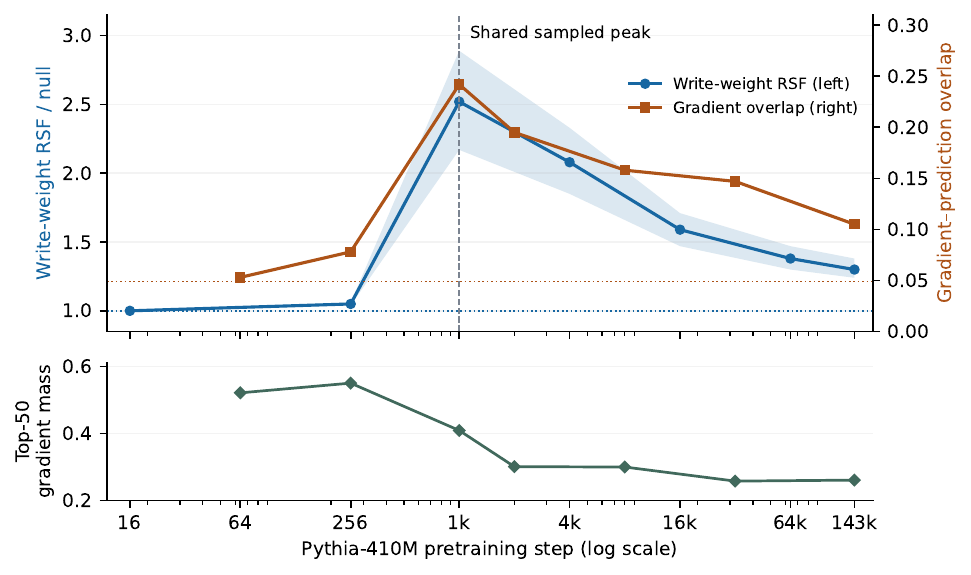}
\caption{Prediction-aligned gradient structure and write-weight concentration develop together early in Pythia-410M pretraining. Blue: within-checkpoint write RSF with a layer and matrix bootstrap interval; orange: gradient-subspace overlap; both peak at step 1000. Lower panel: top-50 gradient covariance mass, which peaks earlier.}
\label{fig:trajectory}
\end{figure}

We follow the formation of the write geometry on the intermediate checkpoints of Pythia-410M, recomputing the prediction projector at each checkpoint so that the trajectory tracks the geometry of the model at that point in training (Figure~\ref{fig:trajectory} and Table~\ref{tab:pythia}). The gradient overlap with $\pred$ starts close to its $0.049$ baseline at step 64, rises to $0.078$ at step 256, peaks at $0.242$, five times the baseline, at step 1000, and remains above baseline at the final checkpoint, while the top-50 gradient mass peaks earlier, at step 256, so concentration and orientation follow different schedules under the same objective.


The write weights follow the gradients, since write RSF is $1.00$ at step 16 and still $1.05$ at step 256, when the gradient overlap has already risen, then peaks at $2.52$ at step 1000, together with the gradient overlap, and declines to $1.30$ by step 143000. The prediction subspace itself moves over the same period, since its overlap with the initial subspace falls to $0.149$ by step 1000 and then drifts only to $0.135$ by step 143000, so most of its departure from initialisation is complete when the write weights peak, and the rise and decay is the same when interval deltas are measured against the prediction projector of the first checkpoint, of the start of each interval or of the final checkpoint (Appendix~\ref{app:gradient}). The write pathway thus acquires its prediction geometry in an early period in which the gradients point into $\pred$, after which the share of write-weight energy in the prediction subspace declines towards an above-random level. OLMo-2-1B and Amber-7B show the same rise and decay, from RSF $1.30$ to $1.03$ and from about $1.6$ to $1.2$ over their runs. Training stage, not the objective, decides the orientation of the write-side factor, and every objective of Section~\ref{sec:notspecific} arrives after that period. The prediction geometry is also specific to this kind of training: at the median layer, intermediate activations of autoregressive transformers place $0.149$--$0.163$ of their energy in the prediction subspace, against $0.010$--$0.029$ for the Dream diffusion language model and $0.010$ for Mamba \citep{mamba}, with a randomly initialised transformer at $0.016$, close to its $k/d$ (Appendix~\ref{app:bridge}).

\section{What remains accessible after alignment?}\label{sec:consequences}
This section asks what the geometry described so far leaves open to direct edits of the write matrices and to adaptation of a single group of matrices.
\subsection{Unembedding directions still act on the write pathway}\label{sec:edit}
If alignment leaves the prediction geometry of the write pathway in place, a write direction derived only from the unembedding should still affect predictions in aligned models. We test exactly this; the experiment does not show how training produced the geometry, and the mechanism of Section~\ref{sec:mechanism} does not rest on it. With normalised unembedding rows $\hat u_i=W_U[i]/\|W_U[i]\|_2$ and a set of targets $\mathcal C$, we form a weighted centroid and edit the write matrices as
\begin{equation}
 \begin{aligned}
 u_{\mathcal C}&=\frac{\sum_{i\in\mathcal C}w_i\hat u_i}{\|\sum_{i\in\mathcal C}w_i\hat u_i\|_2},\\
 W&\leftarrow W-\alpha\, u_{\mathcal C}(u_{\mathcal C}^\top W),\quad W\in\{W_O,W_2\},
 \end{aligned}
 \label{eq:edit}
\end{equation}
a change of rank at most one per matrix, applied at every layer with $\alpha=0.5$ and evaluated on Wikipedia continuations about capitals, continents and chemical elements, with 1,778 prompts where token coverage is complete and 964 for the single-token subset of Mistral (Appendix~\ref{app:controls}).

\begin{table}[t]
\centering
\small
\begin{tabular}{llccc}
\toprule
Model & Stage & $\Delta\log p_{\mathrm{target}}$ & $\Delta\log p_{\mathrm{cont}}$ & PPL\\
\midrule
Llama-3.2-3B & Base & $-3.01\pm1.26$ & $-1.70$ & $1.05$\\
 & SFT & $-2.13\pm1.15$ & $-0.98$ & $1.05$\\
Llama-3-8B & Base & $-3.43\pm1.29$ & $-1.10$ & $1.01$\\
 & SFT & $-2.67\pm1.59$ & $-0.30$ & $1.01$\\
Tulu-3-8B & SFT & $-2.65$ & $-0.61$ & $1.01$\\
 & DPO & $-1.94\pm1.46$ & $-0.04$ & $1.00$\\
Qwen2.5-3B & SFT & $-2.29\pm1.34$ & $-1.23$ & $1.08$\\
Qwen2.5-7B & SFT & $-2.81\pm1.44$ & $-0.67$ & $1.11$\\
Mistral-7B & SFT & $-0.92\pm1.16$ & $+0.20$ & $1.00$\\
OLMo-2-7B & DPO & $\mathbf{-0.03\pm0.21}$ & --- & ---\\
\bottomrule
\end{tabular}
\caption{Rank-1 write edits at $\alpha=0.5$ in the static cohort. Target and continuation changes are in nats, as means ($\pm$ standard deviation across prompts where available); PPL is the off-concept perplexity ratio, and a dash denotes an unavailable measurement.}
\label{tab:edit}
\end{table}

The edit suppresses targets by $3.01$ and $3.43$ nats in the two base models and by $1.94$ to $2.81$ nats in six of the eight aligned checkpoints (Table~\ref{tab:edit}), so a direction computed from the unembedding alone still acts on the write pathway after alignment, with somewhat smaller effects than in the base models and an off-concept perplexity cost of 0 to 11 percent. Random unit directions and directions in the bottom-50 right-singular subspace of $W_U$ at the same nominal strength change target log-probability by at most $0.02$ nats on average (Table~\ref{tab:controls}), but they do not match the realised norm of the edit, so we repeat the edit on the controlled Pythia-410M checkpoints, the anchor and its three branches, with eight draws per control family matched in norm to the centroid edit matrix by matrix, on 60 prompts covering 30 single-token targets, with retention measured on held-out WikiText windows (Appendix~\ref{app:controls}). At $\alpha=0.5$ the centroid lowers target log-probability by $3.8$ to $4.1$ nats on the four checkpoints, while norm-matched random directions change it by $0.02$ to $0.03$ nats on average, random directions inside the prediction subspace by at most $0.02$, and bottom-spectrum directions by $0.04$ to $0.15$, with single draws reaching $1.0$ nats (Table~\ref{tab:normmatched}), so the effect is target-directed beyond what the magnitude of the edit produces, although bottom-spectrum directions are not inert. On this small model the edit has a large retention cost, since the full centroid raises retention perplexity by 31 to 65 percent at $\alpha=0.5$, and by 7 to 17 percent at $\alpha=0.25$, where it still removes $1.6$ to $1.7$ nats. Splitting the centroid into its components inside and outside the leading prediction subspace, which hold 48 and 52 percent of its energy, locates the cost: the inside component suppresses targets by $3.0$ to $3.6$ nats but doubles or triples retention perplexity, while the outside component suppresses them by $1.3$ to $1.5$ nats at no measurable cost. A direction built from the unembedding is therefore enriched in the prediction subspace without being contained in it, and the part of its effect that costs little retention lies outside, which matches the finding of Section~\ref{sec:dynamics} that the structured signal the write pathway receives lies outside $\pred$.

\begin{table}[t]
\centering
\small
\begin{tabular}{lcc}
\toprule
Direction & $\Delta\log p_{\mathrm{target}}$ & PPL ratio\\
\midrule
Full centroid & $-3.78$ to $-4.06$ & $1.31$ to $1.65$\\
Centroid inside $\pred$ & $-3.00$ to $-3.63$ & $1.93$ to $3.03$\\
Centroid outside $\pred$ & $-1.33$ to $-1.54$ & $1.00$\\
Random (matched) & $-0.02$ to $-0.03$ & $1.00$ to $1.01$\\
Random in $\pred$ (matched) & $-0.02$ to $+0.01$ & $1.00$ to $1.02$\\
Random bottom-50 (matched) & $-0.04$ to $-0.15$ & $1.02$ to $1.05$\\
\bottomrule
\end{tabular}
\caption{Norm-matched write edits at $\alpha=0.5$ on the controlled Pythia-410M checkpoints, as ranges over the anchor and its three branches: mean target change in nats, averaged over eight draws for random families, and retention perplexity ratio (Table~\ref{tab:normmatched-full}).}
\label{tab:normmatched}
\end{table}

Two aligned checkpoints respond weakly. Mistral single-tokenises only 19 of the 32 targets, and edits along its single-token directions reduce each target by $1.5$ to $5.4$ nats (Table~\ref{tab:mistral}), so its reduced coverage lowers the aggregate effect, whereas OLMo-2-7B DPO has full coverage and does not respond ($-0.03\pm0.21$ nats). A high stable rank of $\Delta W_U$ accompanies the non-responding checkpoints, but Tulu responds at the largest stable rank in the sample (Table~\ref{tab:srank}), so in this one checkpoint the centroid fails for a reason the unembedding delta alone does not reveal.

\subsection{Every tier can adapt}\label{sec:lora}
The signature could also be read as showing that later training happens in the read pathway because only the read pathway can carry it. If it instead reflects where the gradients place their updates, restricting adaptation to a single tier should still work. We compare read-only ($W_Q,W_K$), write-only ($W_O,W_2$), internal-only ($W_V,W_1$) and all-pathway LoRA adapters on held-out SFT loss, with a common SFT recipe, two seeds, and ranks chosen so that every placement has the same number of trainable parameters (Table~\ref{tab:lora}).

\begin{table}[t]
\centering
\small
\begin{tabular}{lccc}
\toprule
Placement & Loss gain & WikiText PPL & RSF/$\pred$\\
\midrule
Read-only & $0.205$ & $+0.4$ to $+0.5\%$ & $1.04$--$1.07$\\
Write-only & $0.244$ & $-0.3\%$ & $1.14$--$1.15$\\
Internal-only & $0.250$ & $-0.5$ to $-0.6\%$ & $1.04$--$1.07$\\
All pathways & $0.248$ & $-0.5\%$ & $0.98$--$1.17$\\
\bottomrule
\end{tabular}
\caption{Pathway-restricted LoRA under a matched parameter budget, over two seeds: held-out SFT loss gain in nats, WikiText perplexity change, and RSF of the learned update in $\pred$.}
\label{tab:lora}
\end{table}
Read-only adapters recover about 83 percent of the all-pathway loss gain ($0.205/0.248$ nats), write-only and internal-only adapters match it, with $0.244$ and $0.250$ nats, and WikiText perplexity changes by less than one percent in every placement. The write-only updates remain only weakly enriched in $\pred$, with RSF $1.14$--$1.15$, as the natural alignment deltas are, so effective adaptation happens largely outside the leading prediction subspace, as Section~\ref{sec:dynamics} and the decomposition of the edit anticipate. The signature therefore describes where training places its updates when every matrix is free to move, not which matrices can carry them. A further measurement agrees: read-side and write-side edits of equal nominal strength, injected at the same layer along the same direction, change the attention scores of the following layers by nearly the same amount, with read-to-write ratios of $0.93$--$0.97$ in every model and stage (Appendix~\ref{app:propagation}), so neither pathway is privileged in how far a change propagates.

\section{Discussion}\label{sec:discussion}
The measurements describe one mechanism, two stages of training acting on the two sides of the same outer product. On the read side the raw gradient follows the activation spectrum almost exactly, yet the realised deltas of the static cohort give the leading input directions 12 to 29 times less energy than their share of variance, and the controlled branches place this gap in the optimiser rather than in the data, since cancellation between tokens is as strong inside the principal input subspace as outside it and the Adam step keeps about half of the enrichment of the gradient it is computed from. The orientation of an alignment delta is therefore a joint property of the activation statistics, which supply the candidate directions, and of coordinatewise normalisation, which decides how much of each candidate is kept, so the shape of a delta is partly a property of the optimiser rather than of the data.

On the write side, every objective we applied after the prediction geometry had formed sends the write matrices a strongly structured gradient that lies outside the subspace through which the unembedding reads them, and the geometry they do have in that subspace dates from the first thousand steps of pretraining, when the gradients pointed into it. Two results sharpen this statement: the E5-Mistral-7B pair shows that a late objective can still produce substantial write-side prediction enrichment, so where an objective sends its gradients matters and not only when it is applied, and the norm-matched edits show that the direction which still acts on the write pathway after alignment is enriched in the prediction subspace without being contained in it, with the part of its effect that costs no retention lying outside, where the structured write signal of later training also lies. For the view that alignment is shallow \citep{zhou2023lima,lin2024unlocking,qi2024safetyalignmentjusttokens}, the weight-space counterpart is not that alignment touches a special set of matrices, since it changes every matrix we measure and any single group of matrices can carry it, but that it occurs late, so its updates are oriented as every late update's are, along the directions the inputs supply; comparisons between alignment deltas, before merging for instance, should accordingly be made in the principal input subspace of the read pathway.

\section{Conclusion}
The direction of a weight change, measured in the coordinate system of the residual stream, separates the stages of transformer training more sharply than its size or rank, and the two factors of the gradient explain why: read matrices inherit the geometry of the activations, which every objective shares, and write matrices inherit the geometry of the upstream gradients, which point into the prediction subspace only early in pretraining. The realised updates keep part of this geometry, which is enough for unembedding-derived directions to remain effective after alignment and for any single group of matrices to adapt.

\section{Limitations}\label{sec:limitations}
The pretraining trajectory, the late interval and the controlled branches come from one small model, and the branches run 128 steps from a warmed anchor, so they show which objectives produce the signature rather than how it develops over a full alignment run; within them, the continued cross-entropy branch trains on the calibration domain, which may contribute to its larger read enrichment, and the DPO branch does not reach consistent behavioural alignment, so the intervals we report are descriptive. We attribute the larger allocation shortfall of the static deltas to the optimiser effect measured over those 128 steps, without tracking it over their training. The prediction subspace is defined in the basis of the final residual stream, which intermediate layers share only approximately. The norm-matched edits use one seed and 60 prompts, the edits on the larger models use unit directions at a common strength, and the LoRA comparison uses two seeds and SFT loss rather than a behavioural benchmark.

\section{Ethical considerations}
The edits we study can suppress useful information or change safety-relevant behaviour, and since suppressing a continuation does not establish safer generation, our results support the analysis of such edits rather than their deployment. All models and datasets are public.

\newpage
\bibliographystyle{acl_natbib}
\bibliography{references}

@inproceedings{biderman2023pythia,
  title={Pythia: A suite for analyzing large language models across training and scaling},
  author={Biderman, Stella and Schoelkopf, Hailey and Anthony, Quentin Gregory and Bradley, Herbie and O’Brien, Kyle and Hallahan, Eric and Khan, Mohammad Aflah and Purohit, Shivanshu and Prashanth, USVSN Sai and Raff, Edward and others},
  booktitle={International conference on machine learning},
  pages={2397--2430},
  year={2023},
  organization={PMLR}
}

@article{hu2021lora,
  title={Lora: Low-rank adaptation of large language models},
  author={Hu, Edward J and Shen, Yelong and Wallis, Phillip and Allen-Zhu, Zeyuan and Li, Yuanzhi and Wang, Shean and Wang, Lu and Chen, Weizhu},
  journal={arXiv preprint arXiv:2106.09685},
  year={2021}
}

@article{rafailov2023direct,
  title={Direct preference optimization: Your language model is secretly a reward model},
  author={Rafailov, Rafael and Sharma, Archit and Mitchell, Eric and Manning, Christopher D and Ermon, Stefano and Finn, Chelsea},
  journal={Advances in neural information processing systems},
  volume={36},
  pages={53728--53741},
  year={2023}
}

@inproceedings{hong2024orpo,
  title={Orpo: Monolithic preference optimization without reference model},
  author={Hong, Jiwoo and Lee, Noah and Thorne, James},
  booktitle={Proceedings of the 2024 Conference on Empirical Methods in Natural Language Processing},
  pages={11170--11189},
  year={2024}
}

@article{zhou2023lima,
  title={Lima: Less is more for alignment},
  author={Zhou, Chunting and Liu, Pengfei and Xu, Puxin and Iyer, Srinivasan and Sun, Jiao and Mao, Yuning and Ma, Xuezhe and Efrat, Avia and Yu, Ping and Yu, Lili and others},
  journal={Advances in Neural Information Processing Systems},
  volume={36},
  pages={55006--55021},
  year={2023}
}

@inproceedings{lin2024unlocking,
  title={The Unlocking Spell on Base {LLM}s: Rethinking Alignment via In-Context Learning},
  author={Lin, Bill Yuchen and Ravichander, Abhilasha and Lu, Ximing and Dziri, Nouha and Sclar, Melanie and Chandu, Khyathi and Bhagavatula, Chandra and Choi, Yejin},
  booktitle={International Conference on Learning Representations (ICLR)},
  year={2024},
  url={https://openreview.net/forum?id=wxJ0eXwwda}
}

@article{raghavendra2024revisiting,
  title={Revisiting the superficial alignment hypothesis},
  author={Raghavendra, Mohit and Nath, Vaskar and Hendryx, Sean},
  journal={arXiv preprint arXiv:2410.03717},
  year={2024}
}

@misc{qi2024safetyalignmentjusttokens,
      title={Safety Alignment Should Be Made More Than Just a Few Tokens Deep}, 
      author={Xiangyu Qi and Ashwinee Panda and Kaifeng Lyu and Xiao Ma and Subhrajit Roy and Ahmad Beirami and Prateek Mittal and Peter Henderson},
      year={2024},
      eprint={2406.05946},
      archivePrefix={arXiv},
      primaryClass={cs.CR},
      url={https://arxiv.org/abs/2406.05946}, 
}

@inproceedings{aghajanyan2021intrinsic,
  title={Intrinsic dimensionality explains the effectiveness of language model fine-tuning},
  author={Aghajanyan, Armen and Gupta, Sonal and Zettlemoyer, Luke},
  booktitle={Proceedings of the 59th annual meeting of the association for computational linguistics and the 11th international joint conference on natural language processing (volume 1: long papers)},
  pages={7319--7328},
  year={2021}
}

@article{circuits,
  title   = {A Mathematical Framework for Transformer Circuits},
  author  = {Elhage, Nelson and Nanda, Neel and Olsson, Catherine and Henighan, Tom and Joseph, Nicholas and Mann, Ben and Askell, Amanda and Bai, Yuntao and Chen, Anna and Conerly, Tom and DasSarma, Nova and Drain, Dawn and Ganguli, Deep and Hatfield-Dodds, Zac and Hernandez, Danny and Jones, Andy and Kernion, Jackson and Lovitt, Liane and Ndousse, Kamal and Amodei, Dario and Brown, Tom and Clark, Jack and Kaplan, Jared and McCandlish, Sam and Olah, Chris},
  journal = {Transformer Circuits Thread},
  year    = {2021},
  url     = {https://transformer-circuits.pub/2021/framework/index.html}
}

@inproceedings{geva2021transformer,
  title={Transformer feed-forward layers are key-value memories},
  author={Geva, Mor and Schuster, Roei and Berant, Jonathan and Levy, Omer},
  booktitle={Proceedings of the 2021 conference on empirical methods in natural language processing},
  pages={5484--5495},
  year={2021}
}

@article{belrose2023eliciting,
  title={Eliciting latent predictions from transformers with the tuned lens},
  author={Belrose, Nora and Ostrovsky, Igor and McKinney, Lev and Furman, Zach and Smith, Logan and Halawi, Danny and Biderman, Stella and Steinhardt, Jacob},
  journal={arXiv preprint arXiv:2303.08112},
  year={2023}
}

@inproceedings{ethayarajh2019contextual,
  title={How contextual are contextualized word representations? Comparing the geometry of BERT, ELMo, and GPT-2 embeddings},
  author={Ethayarajh, Kawin},
  booktitle={Proceedings of the 2019 conference on empirical methods in natural language processing and the 9th international joint conference on natural language processing (EMNLP-IJCNLP)},
  pages={55--65},
  year={2019}
}

@inproceedings{rome,
  title     = {Locating and Editing Factual Associations in {GPT}},
  author    = {Meng, Kevin and Bau, David and Andonian, Alex and Belinkov, Yonatan},
  booktitle = {Advances in Neural Information Processing Systems},
  volume    = {35},
  year      = {2022}
}

@article{meng2022mass,
  title={Mass-editing memory in a transformer},
  author={Meng, Kevin and Sharma, Arnab Sen and Andonian, Alex and Belinkov, Yonatan and Bau, David},
  journal={arXiv preprint arXiv:2210.07229},
  year={2022}
}

@article{belrose2023leace,
  title={Leace: Perfect linear concept erasure in closed form},
  author={Belrose, Nora and Schneider-Joseph, David and Ravfogel, Shauli and Cotterell, Ryan and Raff, Edward and Biderman, Stella},
  journal={Advances in Neural Information Processing Systems},
  volume={36},
  pages={66044--66063},
  year={2023}
}

@article{arditi2024refusal,
  title={Refusal in language models is mediated by a single direction},
  author={Arditi, Andy and Obeso, Oscar and Syed, Aaquib and Paleka, Daniel and Panickssery, Nina and Gurnee, Wes and Nanda, Neel},
  journal={Advances in Neural Information Processing Systems},
  volume={37},
  pages={136037--136083},
  year={2024}
}

@article{ilharco2022editing,
  title={Editing models with task arithmetic},
  author={Ilharco, Gabriel and Ribeiro, Marco Tulio and Wortsman, Mitchell and Gururangan, Suchin and Schmidt, Ludwig and Hajishirzi, Hannaneh and Farhadi, Ali},
  journal={arXiv preprint arXiv:2212.04089},
  year={2022}
}

@article{yadav2023ties,
  title={Ties-merging: Resolving interference when merging models},
  author={Yadav, Prateek and Tam, Derek and Choshen, Leshem and Raffel, Colin A and Bansal, Mohit},
  journal={Advances in neural information processing systems},
  volume={36},
  pages={7093--7115},
  year={2023}
}

@inproceedings{mamba,
  title     = {Mamba: Linear-Time Sequence Modeling with Selective State Spaces},
  author    = {Gu, Albert and Dao, Tri},
  booktitle = {First Conference on Language Modeling},
  year      = {2024}
}

\clearpage
\appendix
\renewcommand{\thetable}{A\arabic{table}}
\renewcommand{\theequation}{A\arabic{equation}}
\setcounter{table}{0}
\setcounter{equation}{0}

\section{Geometric quantities}\label{app:terms}
The measurements in this paper separate four properties of an update, namely how large it is, how concentrated its spectrum is, which subspace it occupies, and what it does behaviourally. RSF measures the third. It is unchanged when a matrix is multiplied by a nonzero scalar, so a large update can have near-baseline RSF and a small update can have high RSF. Table~\ref{tab:terms} collects the terms used throughout.

\begin{table}[h]
\centering
\small
\begin{tabularx}{\linewidth}{lX}
\toprule
Term & Meaning\\
\midrule
$\pred$ & Rank-50 projector onto the top right-singular vectors of $W_U$.\\
$\inp$ & Layer-specific projector onto the principal subspace of the layer-input activations on a calibration corpus.\\
$\beh$ & Projector onto the principal subspace of final-layer activation differences between matched aligned and base checkpoints.\\
RSF ratio & Projected Frobenius energy fraction divided by $k/d$ (Equation~\ref{eq:rsf}); one is the random-orientation expectation.\\
Top-$k$ spectral mass & Fraction of a covariance trace in its $k$ largest eigenvalues.\\
Top-subspace overlap & Mean squared principal-angle cosine (Equation~\ref{eq:overlap}); random-subspace expectation $k/d$.\\
Direct reads & Query and key matrices, whose input side carries residual-stream coordinates.\\
Internal reads & Value and MLP input (gate and up) matrices, which read residual inputs and affect the residual stream through a subsequent output map.\\
Writes & Attention and MLP output matrices, whose output side carries residual-stream coordinates.\\
Stable rank & $\|\Delta W_U\|_F^2/\|\Delta W_U\|_2^2$.\\
\bottomrule
\end{tabularx}
\caption{Terminology.}
\label{tab:terms}
\end{table}

\section{Checkpoints and cohorts}\label{app:cohorts}
The static analysis uses matched checkpoint differences, $\Delta W=W_{\mathrm{aligned}}-W_{\mathrm{base}}$, and for two-stage pipelines also the isolated second stage, $W_{\mathrm{DPO}}-W_{\mathrm{SFT}}$. Table~\ref{tab:checkpoints} lists the checkpoints and Table~\ref{tab:cohorts} the measurements each cohort supports. The Llama-3-8B DPO and Qwen2.5-7B DPO comparisons are excluded from the static averages, since the former has a near-zero delta in the matrices of Table~\ref{tab:static-full} and the latter is identical to SFT at four decimal places.

\begin{table}[h]
\centering
\footnotesize
\setlength{\tabcolsep}{4pt}
\begin{tabular}{lllll}
\toprule
Group & Base & SFT & DPO & ORPO\\
\midrule
Llama-3B & Llama-3.2-3B & Llama-3.2-3B-Instruct & tanliboy/Llama-3.2-3B-DPO & --\\
Llama-8B & Meta-Llama-3-8B & Meta-Llama-3-8B-Instruct & OpenHermes-DPO & Llama-3-8B-Orpo-v0.1\\
Tulu-8B & Llama-3.1-8B & Tulu-3-8B-SFT & Tulu-3-8B-DPO & --\\
OLMo-7B & OLMo-2-7B & OLMo-2-7B-SFT & OLMo-2-7B-DPO & --\\
Qwen-3B & Qwen2.5-3B & Qwen2.5-3B-Instruct & Qwen2.5-3B-dpo-tuned & Qwen2.5-3B-orpo\\
Qwen-7B & Qwen2.5-7B & Qwen2.5-7B-Instruct & Qwen2.5-7B-DPO-main & --\\
Mistral-7B & Mistral-7B-v0.1 & Mistral-7B-Instruct-v0.1 & zephyr-7b-beta & Mistral-7B-ORPO-beta\\
\bottomrule
\end{tabular}
\caption{Checkpoints used in the static alignment-delta analysis. A dash indicates that no matched checkpoint was used.}
\label{tab:checkpoints}
\end{table}

\begin{table}[h]
\centering
\small
\begin{tabularx}{\linewidth}{lX}
\toprule
Cohort & Measurements\\
\midrule
Static alignment cohort & Base-to-aligned and isolated DPO-minus-SFT deltas for the seven groups of Table~\ref{tab:checkpoints}; rank-1 edits and controls; downstream propagation of read and write edits (Appendix~\ref{app:propagation}).\\
Pythia-410M & Within-checkpoint write-weight trajectory, prediction-subspace trajectory, raw activation spectra at initialisation and at the trained checkpoint, upstream-gradient trajectory; late pretraining interval and controlled branches with gradient, optimiser and norm-matched edit measurements (Appendix~\ref{app:controlled}).\\
OLMo-2-1B & Post-normalisation activation spectra; upstream-gradient spectra under generic-text CE, SFT and DPO; pretraining weight trajectory.\\
Amber-7B & Pretraining weight trajectory.\\
Attention-score cohort & Llama-3.2-1B, Llama-3.2-3B and its instruct model, OLMo-1B, Qwen2.5-3B and its instruct model, Qwen3-0.6B, and five public pretraining ladders (Appendix~\ref{app:selection-details}).\\
Contrastive pairs & MiniLM-L6, MPNet-base, GTE-Qwen2-1.5B, E5-Mistral-7B, with Mistral-7B SFT as a matched comparator (Appendix~\ref{app:objectives}).\\
Dream-7B, Mamba-370M & Activation-level comparison (Appendix~\ref{app:bridge}).\\
Restricted LoRA & Two-seed, matched-budget comparison of adapter placements (Appendix~\ref{app:lora}).\\
\bottomrule
\end{tabularx}
\caption{Cohorts and the measurements each supports.}
\label{tab:cohorts}
\end{table}

\section{Projectors, RSF and the full static-delta table}\label{app:projectors}
All projectors have rank $k=50$ unless stated otherwise. $\pred$ is built from the top right-singular vectors of $W_U$, using sliced truncated SVD for matrices with more than 4000 rows. $\inp$ is estimated at each layer from the vectors that the matrices of a tier multiply, the residual stream after the pre-block normalisation at the input of the attention block for $W_Q$, $W_K$ and $W_V$ and at the input of the MLP block for $W_1$, which are the input factor $A$ of Equation~\ref{eq:gradient} for those matrices; its columns are the leading principal directions of these vectors over the calibration set, base-model activations on WikiText-2 by default. $\beh$ is built from the principal components of final-layer activation differences between the aligned and base checkpoints on the same inputs. Layer-level RSF is the mean over the matrices of a tier, and a checkpoint-level value is the mean over layers; for internal reads, the value of a comparison is the mean of $W_V$ and the gate and up projections.

Table~\ref{tab:static-full} lists the per-model RSF values. Panel A contains base-to-aligned differences, including full two-stage pipelines, and Panel B isolates the second stage. Table~\ref{tab:static} of the main text uses the arithmetic mean and sample standard deviation over the 14 Panel A rows with equal weight per checkpoint.

\begin{table}[h]
\centering
\small
\begin{tabular}{llccccccr}
\toprule
 & & \multicolumn{2}{c}{$\pred$} & \multicolumn{2}{c}{$\inp$} & \multicolumn{2}{c}{$\beh$} & \\
\cmidrule(lr){3-4}\cmidrule(lr){5-6}\cmidrule(lr){7-8}
Model & Stage & Wr & Rd & Wr & Rd & Wr & Rd & Stable rank\\
\midrule
\multicolumn{9}{l}{\emph{A. Aligned minus base}}\\
Llama-3.2-3B & SFT & 1.07 & 1.25 & 1.17 & 2.15 & 1.10 & 1.34 & 6.1\\
 & DPO & 1.21 & 1.42 & 1.45 & 2.67 & 1.18 & 1.40 & 1.4\\
Llama-3-8B & SFT & 1.03 & 1.02 & 1.28 & 3.14 & 1.04 & 1.11 & 6.0\\
 & ORPO & 1.12 & 1.05 & 1.39 & 1.33 & 1.17 & 1.07 & ---\\
Tulu-3-8B & SFT & 1.06 & 1.01 & 1.29 & 2.17 & 1.04 & 1.07 & 12.1\\
 & DPO & 1.06 & 1.02 & 1.30 & 2.18 & 1.04 & 1.07 & 12.1\\
OLMo-2-7B & SFT & 1.11 & 1.07 & 1.33 & 1.71 & 1.11 & 1.07 & 15.4\\
 & DPO & 1.11 & 1.07 & 1.33 & 1.70 & 1.11 & 1.07 & 15.4\\
Qwen2.5-3B & SFT & 1.07 & 1.13 & 1.14 & 1.59 & 1.05 & 1.06 & 1.8\\
 & ORPO & 1.10 & 1.13 & 1.22 & 1.79 & 1.09 & 1.08 & 1.8\\
Qwen2.5-7B & SFT & 1.10 & 1.01 & 1.19 & 1.76 & 1.01 & 1.00 & 2.5\\
Mistral-7B & SFT & 1.13 & 1.07 & 1.26 & 2.59 & 1.10 & 1.09 & 10.8\\
 & DPO & 1.18 & 1.12 & 1.29 & 3.14 & 1.09 & 1.17 & 25.1\\
 & ORPO & 1.03 & 1.02 & 1.25 & 1.22 & 1.12 & 1.00 & 31.9\\
\midrule
\multicolumn{9}{l}{\emph{B. DPO minus SFT}}\\
Tulu-3-8B & DPO/SFT & 1.15 & 1.06 & 1.66 & 2.08 & 1.07 & 1.06 & 163.6\\
OLMo-2-7B & DPO/SFT & 1.29 & 1.29 & 1.70 & 2.67 & 1.18 & 1.21 & 78.1\\
\bottomrule
\end{tabular}
\caption{Layer-averaged RSF ratios of all static comparisons (Wr: writes, Rd: direct reads). Panel A contains base-to-aligned differences, Panel B the isolated second stage. Stable rank is that of $\Delta W_U$; a dash marks an effectively zero $\Delta W_U$. Table~\ref{tab:static} averages the 14 rows of Panel A.}
\label{tab:static-full}
\end{table}

\paragraph{Internal reads}
Internal reads share residual-stream inputs with direct reads, but $W_V$ acts on the residual stream through $W_O$, and the MLP input maps act through a nonlinearity and $W_2$. Table~\ref{tab:internal} records their orientation. Input-subspace ratios lie mostly between 1.1 and 1.6, with the largest values in Llama-3.2-3B DPO, prediction-subspace ratios stay close to one, and the gate projection exceeds the up projection in $\inp$ in every row that reports both, except Llama-3-8B ORPO and Mistral-7B ORPO.

\begin{table}[h]
\centering
\small
\begin{tabular}{llcccc}
\toprule
 & & \multicolumn{2}{c}{$\pred$} & \multicolumn{2}{c}{$\inp$}\\
\cmidrule(lr){3-4}\cmidrule(lr){5-6}
Model & Stage & $W_V$ & $W_1^{\mathrm{gate/up}}$ & $W_V$ & $W_1^{\mathrm{gate/up}}$\\
\midrule
Llama-3.2-3B & SFT & 1.02 & 1.08/1.08 & 1.15 & 1.31/1.25\\
 & DPO & 1.25 & 1.23/1.22 & 2.07 & 1.98/1.91\\
Llama-3-8B & SFT & 0.98 & 1.02/1.01 & 1.37 & 1.46/1.37\\
 & DPO & 1.09 & --- & 1.60 & ---\\
 & ORPO & 1.06 & 1.02/1.04 & 1.14 & 1.09/1.11\\
Tulu-3-8B & SFT & 0.98 & 1.03/1.01 & 1.48 & 1.51/1.43\\
 & DPO & 0.98 & 1.03/1.02 & 1.48 & 1.51/1.43\\
Mistral-7B & SFT & 1.04 & 1.11/1.09 & 1.31 & 1.40/1.34\\
 & DPO & 1.10 & 1.11/1.10 & 1.39 & 1.44/1.35\\
 & ORPO & 1.04 & 1.03/1.04 & 1.40 & 1.17/1.25\\
Qwen2.5-3B & SFT & 1.07 & 1.07/1.07 & 1.12 & 1.19/1.14\\
 & ORPO & 1.07 & 1.06/1.06 & 1.23 & 1.29/1.23\\
Qwen2.5-7B & SFT & 1.01 & 1.03/1.03 & 1.27 & 1.31/1.24\\
 & DPO & 1.01 & 1.03/1.03 & 1.27 & 1.31/1.24\\
\bottomrule
\end{tabular}
\caption{Internal-read RSF ratios; $W_1$ entries are gate/up values. Base-referenced DPO rows of two-stage pipelines are dominated by the SFT component, and the Llama-3-8B DPO MLP entry is unavailable.}
\label{tab:internal}
\end{table}

\subsection{Statistical model}\label{app:stats}
For a model-stage, layer, tier and projector observation, let $y_i=\log\rsf_i$. We fit
\begin{equation}
 y_i=\beta_0+\beta_1\mathrm{Diag}_i+\beta_2\mathrm{Pathway}_i+\beta_3(\mathrm{Diag}_i\times\mathrm{Pathway}_i)+b_{\mathrm{base}[i]}+\epsilon_i,
 \label{eq:mixed}
\end{equation}
where Diag marks the pairing the mechanism assigns, $\inp$ for reads and $\pred$ for writes, and $b_{\mathrm{base}}$ is a random intercept per base model, with Wald tests for the coefficients. The read contrast is $\beta=0.98$ ($z=5.89$, $p=4\times10^{-9}$) and the write contrast $\beta=-0.18$ ($z=-6.81$, $p=10^{-11}$), so writes are less enriched in $\pred$ than in the other references taken together. The architecture-level test reduces each of the seven base models to the mean read contrast over its comparisons and applies an exact one-sided Wilcoxon signed-rank test, which gives $p=0.008$, the smallest value attainable with seven units, since the contrast is positive in all of them. Because the seven base models contain dependent layer observations, the effect sizes and their consistency across families carry more weight than the layer-level $p$-values.

\section{Projector and calibration robustness}\label{app:robustness}
Tables~\ref{tab:rank}--\ref{tab:source} vary the projector rank, the calibration corpus, and the model whose activations define $\inp$. The read signature stays above one in every case. Its magnitude changes smoothly with rank, it is largest for the two instruction corpora, OASST1 and Alpaca, in all three models, and it changes by less than $0.07$ when the projector is built from aligned rather than base activations, with bootstrap intervals for the difference of $[-0.087,-0.041]$ for Llama and $[+0.015,+0.073]$ for Mistral. The rank comparison uses the projectors of the default run, so its $k=50$ column repeats Table~\ref{tab:static-full}. The calibration-corpus comparison and the projector-source comparison, whose base-projector values are also the SFT values of Table~\ref{tab:mistral-objective}, were computed in runs separate from the default one, which is why the WikiText-2 read value of one checkpoint differs between Tables~\ref{tab:static-full}, \ref{tab:calibration} and~\ref{tab:source}; each comparison is to be read within its table.

\begin{table}[h]
\centering
\small
\begin{tabular}{lccc}
\toprule
Model & $k=32$ & $k=50$ & $k=64$\\
\midrule
Llama-3.2-3B SFT & 2.30 [2.15, 2.46] & 2.15 [2.01, 2.30] & 2.05 [1.92, 2.18]\\
Mistral-7B SFT & 2.78 [2.51, 3.07] & 2.59 [2.34, 2.85] & 2.45 [2.21, 2.69]\\
Qwen2.5-7B SFT & 1.91 [1.74, 2.09] & 1.76 [1.61, 1.93] & 1.65 [1.51, 1.81]\\
\bottomrule
\end{tabular}
\caption{Read/$\inp$ RSF under three projector ranks, with bootstrap intervals over layers.}
\label{tab:rank}
\end{table}

\begin{table}[h]
\centering
\small
\begin{tabular}{lcccc}
\toprule
Model & WikiText-2 & C4 & OASST1 & Alpaca\\
\midrule
Llama-3-8B & 3.36 & 3.83 & 4.21 & 4.01\\
Mistral-7B & 2.22 & 2.95 & 3.26 & 3.17\\
Qwen2.5-7B & 1.82 & 1.87 & 2.12 & 2.08\\
\bottomrule
\end{tabular}
\caption{Read/$\inp$ RSF of the SFT deltas across calibration corpora.}
\label{tab:calibration}
\end{table}

\begin{table}[h]
\centering
\small
\begin{tabular}{lcccc}
\toprule
Model & Base projector & Aligned projector & Difference & Mean cosine\\
\midrule
Llama-3.2-3B & 1.808 [1.755, 1.862] & 1.744 [1.697, 1.792] & $-0.064$ [$-0.087$, $-0.041$] & 0.889\\
Mistral-7B & 1.769 [1.648, 1.927] & 1.808 [1.668, 1.995] & $+0.038$ [$+0.015$, $+0.073$] & 0.886\\
\bottomrule
\end{tabular}
\caption{Read/$\inp$ RSF with projectors estimated from base and from aligned activations, with 5,000-resample layer-bootstrap intervals. The last column is the unsquared mean principal-angle cosine between the two projectors.}
\label{tab:source}
\end{table}

\section{Late pretraining interval and controlled branches}\label{app:controlled}
The late interval compares the Pythia-410M checkpoints at steps 128,000 and 143,000, with input projectors built from the post-normalisation query, key and value inputs of the step-128,000 checkpoint on 64 WikiText-2 sequences of 512 tokens, centred by default, with eight random projectors of the same rank as a null (mean RSF $1.003$) and 5,000-resample layer bootstraps. Table~\ref{tab:late} gives the interval delta's read enrichment under rank and centering changes; the enrichment is present at every rank and under both conventions, and the raw projected fraction grows with rank as the subspace grows.

\begin{table}[h]
\centering
\small
\begin{tabular}{lcccccc}
\toprule
Centering & $k$ & Read & $W_Q$ & $W_K$ & Energy & Null\\
\midrule
Centred & 32 & 4.44 & 7.16 & 1.71 & 0.139 & 1.004\\
 & 50 & 3.30 & 5.04 & 1.57 & 0.161 & 1.003\\
 & 64 & 2.84 & 4.19 & 1.50 & 0.178 & 1.005\\
Uncentred & 32 & 4.82 & 7.82 & 1.82 & 0.151 & 1.004\\
 & 50 & 3.54 & 5.44 & 1.64 & 0.173 & 1.003\\
 & 64 & 3.03 & 4.50 & 1.56 & 0.189 & 1.005\\
\bottomrule
\end{tabular}
\caption{Read/$\inp$ RSF of the Pythia-410M interval delta from step 128,000 to 143,000 under rank and centering changes, over 24 layers; Energy is the raw projected fraction and Null the mean RSF against eight random projectors.}
\label{tab:late}
\end{table}

The controlled branches start from the step-143,000 checkpoint. For each of three seeds (17, 23, 41), 64 SFT steps on the chosen responses of \nolinkurl{trl-lib/ultrafeedback_binarized} produce a warmed anchor, and three branches of 128 steps each are trained from that anchor, continued cross-entropy on WikiText-2, SFT on the chosen responses, and DPO against the anchor as reference with $\beta=0.1$ and response-summed log-probabilities. All branches use a fresh AdamW with learning rate $5\times10^{-6}$, $\beta=(0.9,0.99)$, $\epsilon=10^{-8}$, no weight decay, gradient clipping at 1, batch size 2 and sequence length 256, in FP32 with FP64 geometry; seeds vary the data order and the warmup. Input tokens per branch are 65,536 for cross-entropy, 52,518--54,374 for SFT and 105,268--109,666 for DPO, which evaluates both responses. Projectors for the branches use the post-normalisation inputs of each anchor on 64 WikiText-2 sequences of 256 tokens. Held-out evaluation uses 128 preference pairs, shared across seeds, with chosen-response negative log-likelihood, WikiText-2 negative log-likelihood, and the fraction of pairs on which the log-probability margin between chosen and rejected responses improves over the anchor (Table~\ref{tab:controlled-eval}). The continued cross-entropy branch lowers WikiText-2 loss and raises the chosen-response loss, SFT lowers the chosen-response loss, and DPO raises both while improving the preference margin on about half of the pairs, so the three branches move the model in different directions while, as Table~\ref{tab:controlled} shows, producing the same orientation signature.

\begin{table}[h]
\centering
\small
\begin{tabular}{lccc}
\toprule
Branch & Chosen NLL & WikiText-2 NLL & Relative preference\\
\midrule
Anchor & 1.901--1.916 & 3.286--3.401 & ---\\
Text CE & 2.085--2.099 & 2.771--2.784 & 0.430--0.453\\
SFT & 1.864--1.875 & 3.346--3.377 & 0.516--0.609\\
DPO & 2.116--2.262 & 3.513--3.566 & 0.461--0.555\\
\bottomrule
\end{tabular}
\caption{Held-out evaluation of the controlled branches on 128 preference pairs, as ranges over three seeds: chosen-response and WikiText-2 negative log-likelihood, and the fraction of pairs whose chosen-minus-rejected margin improves over the anchor.}
\label{tab:controlled-eval}
\end{table}

\section{Gradient measurements and pretraining trajectories}\label{app:gradient}
\paragraph{Upstream gradients at write outputs}
We register backward hooks at the outputs of $W_O$ and $W_2$ and keep the residual-dimensional gradient vector of every token before any aggregation, without applying a parameter update. In OLMo-2-1B each measured cell contains approximately 15,000--123,000 token gradients; since tokens within a sequence are dependent, these counts describe the size of the collection rather than an independent sample size. $C_g$ is the covariance of the collected vectors, and Equation~\ref{eq:gradientstats} gives the two statistics of Table~\ref{tab:grads}. For DPO, every token output receives a gradient that carries the sequence-level weight of the objective. In the controlled branches, write-side upstream gradients are collected at the same outputs at steps 1, 16, 64 and 128 and at layers 0, 12 and 23, over every non-padding position of the probe batch, prompt positions included: 512 token gradients per cell under continued cross-entropy, where both sequences are full, 249--512 under SFT, and 481--1024 under DPO, whose batch encodes both responses of each pair. Their covariance is centred; the uncentred second-moment matrix gives the same $m_k$ and $o_k$ to two decimals.

\paragraph{Read-side diagnostics in the controlled branches}
At the same probe points, for $W_Q$ and $W_K$, we keep the token inputs and upstream gradients of the raw gradient and reconstruct it from its token outer products, with a largest relative error of $7\times10^{-7}$. The shuffled statistic recomputes $GA^\top$ after permuting the token order of $A$, averaged over three permutations; the coherence inside a read-side projector $P$ is $\|\sum_t \delta_t a_t^\top P\|_F/\sum_t\|\delta_t a_t^\top P\|_F$, where $P$ acts on the input side of the outer product as in Equation~\ref{eq:rsf}, with the complement $I-P$ for the outside value; the Adam step is the difference of the weights across the step; the cumulative delta is the difference from the anchor; and the gradient-sum proxy accumulates the clipped gradients scaled by $-\eta$ along the same trajectory, so it is not a separately trained run. Table~\ref{tab:diagnostics} of the main text averages these quantities over seeds, probe steps, layers and the two matrices.

\paragraph{Pythia trajectory}
In the Pythia-410M trajectory, $\pred^{(t)}$ is recomputed at every checkpoint. Table~\ref{tab:pythia} gives the gradient statistics and the write-weight trajectory plotted in Figure~\ref{fig:trajectory}, the latter measured against the prediction projector of the same checkpoint with intervals that resample layers and write matrices, together with the overlap between the prediction subspace at each step and the initial one.

\begin{table}[h]
\centering
\small
\begin{tabular}{rcccc}
\toprule
 & \multicolumn{2}{c}{Gradient} & \multicolumn{2}{c}{Weights and $\pred$}\\
\cmidrule(lr){2-3}\cmidrule(lr){4-5}
Step & $m_k$ & $o_k$ & Write RSF [95\% interval] & Overlap with initial $\pred$\\
\midrule
16 & -- & -- & 1.00 [1.00, 1.00] & 1.000\\
64 & 0.521 & 0.053 & -- & --\\
256 & $\mathbf{0.550}$ & 0.078 & 1.05 [1.04, 1.06] & 0.721\\
1000 & 0.409 & $\mathbf{0.242}$ & $\mathbf{2.52}$ [2.17, 2.89] & 0.149\\
2000 & 0.301 & 0.195 & -- & --\\
4000 & -- & -- & 2.08 [1.85, 2.33] & 0.141\\
8000 & 0.300 & 0.158 & -- & --\\
16000 & -- & -- & 1.59 [1.47, 1.71] & 0.138\\
32000 & 0.258 & 0.147 & -- & --\\
64000 & -- & -- & 1.38 [1.30, 1.47] & 0.135\\
143000 & 0.261 & 0.105 & 1.30 [1.24, 1.38] & 0.135\\
\bottomrule
\end{tabular}
\caption{Pythia-410M pretraining trajectory: gradient statistics of Equation~\ref{eq:gradientstats}, write RSF against $\pred^{(t)}$ of the same checkpoint with intervals over layers and write matrices, and the overlap of $\pred^{(t)}$ with $\pred$ at step 16. Baselines are $50/1024\simeq0.049$, and one for RSF; a dash marks an unmeasured checkpoint.}
\label{tab:pythia}
\end{table}

\paragraph{Stagewise comparison}
For a training interval $[t_j,t_{j+1}]$, the interval delta $D_j=W^{(t_{j+1})}-W^{(t_j)}$ can be measured against the prediction projector at the first checkpoint, at the start of the interval, $t_j$, or at the final checkpoint, which separates the movement of the weights from the movement of the reference subspace. The rise and decay of the within-checkpoint trajectory is the same under all three choices.

\section{Allocation ratios}\label{app:allocation}
Equation~\ref{eq:allocation} compares the delta energy in each activation eigendirection with that direction's share of the activation trace, and it has two aggregates,
\begin{equation}
 \bar c_k=\frac1k\sum_{i=1}^k c_i,\qquad c_{\mathrm{mass}}=\frac{f(\Delta W,\inp)}{\rho_k}=\frac{\sum_{i=1}^k\lambda_ic_i}{\sum_{i=1}^k\lambda_i},\qquad \rho_k=\frac{\sum_{i=1}^k\lambda_i}{\tr(\Sigma_a)},
 \label{eq:cmass}
\end{equation}
the first weighting directions equally and the second by activation variance. Table~\ref{tab:allocation} gives both at layer $\lfloor2L/3\rfloor$, with $\Sigma_a$ estimated from 16 WikiText-2 sequences and tail directions with near-zero eigenvalues excluded. Both are well below one while the raw fraction stays above $k/d$, so the delta occupies the principal input subspace above chance while allocating far less energy to it than activation variance would. In the controlled branches, the same ratio $c_{\mathrm{mass}}$ is computed for the raw gradient and for the Adam step at the probed layers, which gives the values $1.01$ and $0.64$ under continued cross-entropy and $0.71$ and $0.31$ under SFT and DPO quoted in Section~\ref{sec:optimiser}.

\begin{table}[h]
\centering
\small
\begin{tabular}{llcccc}
\toprule
Model & Stage & $\bar c_k$ & $\rho_k$ & $f(\Delta W,\inp)$ & $c_{\mathrm{mass}}$\\
\midrule
Llama-3.2-3B & SFT & 0.083 & 0.477 & 0.027 & 0.056\\
 & DPO & 0.114 & 0.477 & 0.040 & 0.084\\
Llama-3-8B & SFT & 0.096 & 0.442 & 0.031 & 0.070\\
Qwen2.5-3B & SFT & 0.156 & 0.610 & 0.038 & 0.062\\
Qwen2.5-7B & SFT & 0.104 & 0.427 & 0.024 & 0.056\\
Mistral-7B & SFT & 0.053 & 0.469 & 0.016 & 0.034\\
\bottomrule
\end{tabular}
\caption{Read-delta allocation at layer $\lfloor 2L/3\rfloor$ (Equation~\ref{eq:cmass}). $f(\Delta W,\inp)$ is the raw projected fraction, not the normalised RSF.}
\label{tab:allocation}
\end{table}

\section{Edits and controls}\label{app:controls}
\paragraph{Edit and controls in the static cohort}
All edits act on $W_O$ and $W_2$ at every layer with $\alpha=0.5$. The centroid direction follows Equation~\ref{eq:edit}, random directions are standard Gaussian vectors in $\R^d$ normalised to unit length, and bottom-spectrum directions are unit vectors in the span of the 50 right-singular vectors of $W_U$ with the smallest singular values, drawn with standard-normal coefficients. The natural-continuation targets are capitals, continents and chemical elements. Table~\ref{tab:controls} gives the per-model results.

\begin{table}[h]
\centering
\small
\begin{tabular}{lrccc}
\toprule
Model/stage & Prompts & Centroid & Random & Bottom spectrum\\
\midrule
Llama-3.2-3B/base & 1778 & $-3.01\pm1.26$ & $-0.00\pm0.05$ & $-0.01\pm0.05$\\
Llama-3.2-3B/SFT & 1778 & $-2.13\pm1.15$ & $+0.00\pm0.04$ & $+0.00\pm0.05$\\
Llama-3.2-3B/DPO & 1778 & $-2.51\pm1.44$ & $+0.00\pm0.04$ & $+0.01\pm0.04$\\
Llama-3-8B/base & 1778 & $-3.43\pm1.29$ & $+0.00\pm0.03$ & $+0.01\pm0.04$\\
Llama-3-8B/SFT & 1778 & $-2.67\pm1.59$ & $+0.01\pm0.04$ & $-0.00\pm0.04$\\
Tulu-8B/DPO & 1778 & $-1.94\pm1.46$ & $+0.00\pm0.04$ & $+0.01\pm0.04$\\
Qwen2.5-3B/SFT & 1778 & $-2.29\pm1.34$ & $+0.02\pm0.08$ & $-0.02\pm0.15$\\
Qwen2.5-7B/SFT & 1778 & $-2.81\pm1.44$ & $-0.01\pm0.05$ & $-0.01\pm0.07$\\
Mistral-7B/SFT & 964 & $-0.92\pm1.16$ & $+0.00\pm0.02$ & $-0.00\pm0.02$\\
OLMo-2-7B/DPO & 1778 & $-0.03\pm0.21$ & $+0.00\pm0.04$ & $-0.00\pm0.04$\\
\bottomrule
\end{tabular}
\caption{Target log-probability change in nats under centroid, random and bottom-spectrum edits at unit strength, mean $\pm$ standard deviation across prompts.}
\label{tab:controls}
\end{table}

\paragraph{Edit norm}
For a unit direction $u$ the realised edit norm is
\begin{equation}
 \|\alpha uu^\top W\|_F=|\alpha|\,\|u^\top W\|_2,
 \label{eq:editnorm}
\end{equation}
so a control edit $-\alpha_r u_ru_r^\top W$ has the same Frobenius norm as the centroid edit $-\alpha u_cu_c^\top W$ when $\alpha_r=\alpha\,\|u_c^\top W\|_2/\|u_r^\top W\|_2$, or, when matching jointly across layers and matrices, when the corresponding sums of squared norms are equal. The controls of Table~\ref{tab:controls} use unit directions at the common strength $\alpha$; the norm-matched protocol below applies Equation~\ref{eq:editnorm} matrix by matrix.

\paragraph{Tokenizer coverage}
The centroid requires single-token targets, since a multi-token target contributes only the unembedding of its first piece. The Llama, Tulu and Qwen tokenizers single-tokenise all 32 targets of the coverage diagnostic, while Mistral single-tokenises 19, so Mistral is evaluated on its single-token subset. With the centroid built from each single-target direction, Mistral-7B SFT gives the per-target changes of Table~\ref{tab:mistral}, between $-1.53$ and $-5.39$ nats, in line with the Llama and Qwen results.

\begin{table}[h]
\centering
\small
\begin{tabular}{lrlrlr}
\toprule
Target & $\Delta\log p$ & Target & $\Delta\log p$ & Target & $\Delta\log p$\\
\midrule
Paris & $-2.07$ & Europe & $-3.38$ & carbon & $-3.40$\\
Rome & $-2.18$ & Asia & $-4.84$ & oxygen & $-4.30$\\
Madrid & $-1.89$ & Africa & $-3.49$ & iron & $-2.82$\\
Berlin & $-2.45$ & & & copper & $-2.36$\\
Vienna & $-2.26$ & & & silver & $-2.33$\\
Dublin & $-1.53$ & & & tin & $-5.39$\\
Tokyo & $-2.23$ & & & gold & $-2.67$\\
Beijing & $-2.17$ & & & lead & $-2.42$\\
\bottomrule
\end{tabular}
\caption{Per-target suppression in Mistral-7B SFT on its single-token subset, in nats, all layers edited at $\alpha=0.5$.}
\label{tab:mistral}
\end{table}

\paragraph{Stable rank}
The stable rank of the unembedding delta is
\begin{equation}
 \srank(\Delta W_U)=\frac{\|\Delta W_U\|_F^2}{\|\Delta W_U\|_2^2}.
 \label{eq:stablerank}
\end{equation}
Table~\ref{tab:srank} pairs it with the suppression magnitude. Checkpoints with low stable rank respond strongly, and the non-responding DPO and ORPO checkpoints of Mistral and the isolated OLMo-2 DPO delta have high stable rank, but Tulu still gives $1.94$ nats at the largest stable rank in the sample, $163.6$, so stable rank is associated with the edit's success without setting a threshold for it. Stable rank is scale-invariant and depends on the reference checkpoint, measuring spectral spread rather than the magnitude of reorganisation.

\begin{table}[h]
\centering
\small
\begin{tabular}{llrr}
\toprule
Model & Delta reference & Stable rank & Suppression (nats)\\
\midrule
Llama-3.2-3B & DPO/base & 1.4 & 2.51\\
Qwen2.5-3B & SFT/base & 1.8 & 2.29\\
Qwen2.5-3B & ORPO/base & 1.8 & 2.43\\
Qwen2.5-7B & SFT/base & 2.5 & 2.81\\
Llama-3-8B & SFT/base & 6.0 & 2.67\\
Llama-3.2-3B & SFT/base & 6.1 & 2.13\\
Mistral-7B & SFT/base & 10.8 & 0.92\\
Tulu-3-8B & DPO/base & 12.1 & 1.94\\
OLMo-2-7B & SFT/base & 15.4 & 2.65\\
Mistral-7B & DPO/base & 25.1 & $\approx0$\\
Mistral-7B & ORPO/base & 31.9 & $\approx0$\\
OLMo-2-7B & DPO/SFT & 78.1 & 0.03\\
Tulu-3-8B & DPO/SFT & 163.6 & 1.94\\
\bottomrule
\end{tabular}
\caption{Stable rank of $\Delta W_U$ and centroid suppression in nats, ordered by stable rank. Checkpoints appear once per delta reference.}
\label{tab:srank}
\end{table}

\paragraph{Norm-matched edits on the controlled checkpoints}
The norm-matched edits of Section~\ref{sec:edit} act on every $W_O$ and $W_2$ of the seed-17 anchor and branches of Appendix~\ref{app:controlled}, with rank-50 projectors and $\alpha\in\{0.25,0.5\}$. The centroid follows Equation~\ref{eq:edit}; its inside and outside components are its projections onto $\pred$ and onto the complement, each renormalised to a unit direction, and the centroid holds $0.475$--$0.477$ of its energy in $\pred$ on the four checkpoints. Control families are random ambient directions, random directions in $\pred$, and random directions in the span of the 50 right-singular vectors of $W_U$ with the smallest singular values, eight draws each, with $\alpha_r$ chosen per matrix by Equation~\ref{eq:editnorm} so that each control edit has the Frobenius norm of the centroid edit of that matrix, to a largest relative error of $2\times10^{-7}$; the equal-coefficient convention gives the same ordering. Prompts are 64 natural-language and question-form prompts for capitals, continents and chemical elements, of which 60 have a single-token target under the Pythia tokenizer, covering 30 targets with two prompts each. Target changes are equal-target means with 5,000-resample bootstrap intervals over the 30 target means. Retention is the perplexity ratio on 16 held-out WikiText-2 windows of 128 tokens that exclude the exact target token ids. Table~\ref{tab:normmatched-full} gives the full results at $\alpha=0.5$; at $\alpha=0.25$ the full centroid gives $-1.65$, $-1.62$, $-1.69$ and $-1.73$ nats on the anchor and the three branches with perplexity ratios $1.167$, $1.066$, $1.173$ and $1.172$. The ordering is the same on all four checkpoints: the inside component carries most of the suppression and all of the retention cost, the outside component suppresses at no measurable cost, and the three norm-matched control families stay within a few hundredths of a nat on average, with bottom-spectrum draws the most variable.

\begin{table}[h]
\centering
\small
\begin{tabular}{llccc}
\toprule
Checkpoint & Direction & $\Delta\log p$ & 95\% interval / SD & PPL ratio\\
\midrule
Anchor & Full centroid & $-3.937$ & $[-4.1,\,-3.7]$ & 1.597\\
 & Inside $\pred$ & $-3.410$ & $[-3.7,\,-3.0]$ & 2.951\\
 & Outside $\pred$ & $-1.359$ & $[-1.5,\,-1.1]$ & 1.001\\
 & Random & $-0.031$ & 0.062 & 1.003\\
 & Random in $\pred$ & $-0.020$ & 0.082 & 1.015\\
 & Random bottom-50 & $-0.101$ & 0.278 & 1.022\\
Text CE & Full centroid & $-3.783$ & $[-3.9,\,-3.5]$ & 1.309\\
 & Inside $\pred$ & $-3.004$ & $[-3.2,\,-2.7]$ & 1.932\\
 & Outside $\pred$ & $-1.540$ & $[-1.7,\,-1.3]$ & 1.003\\
 & Random & $-0.025$ & 0.050 & 1.002\\
 & Random in $\pred$ & $-0.006$ & 0.051 & 1.008\\
 & Random bottom-50 & $-0.143$ & 0.238 & 1.045\\
SFT & Full centroid & $-3.953$ & $[-4.1,\,-3.7]$ & 1.595\\
 & Inside $\pred$ & $-3.210$ & $[-3.5,\,-2.9]$ & 2.787\\
 & Outside $\pred$ & $-1.498$ & $[-1.7,\,-1.2]$ & 1.001\\
 & Random & $-0.017$ & 0.054 & 1.004\\
 & Random in $\pred$ & $+0.006$ & 0.087 & 1.019\\
 & Random bottom-50 & $-0.150$ & 0.345 & 1.022\\
DPO & Full centroid & $-4.059$ & $[-4.2,\,-3.8]$ & 1.649\\
 & Inside $\pred$ & $-3.630$ & $[-3.9,\,-3.2]$ & 3.034\\
 & Outside $\pred$ & $-1.328$ & $[-1.5,\,-1.1]$ & 1.002\\
 & Random & $-0.020$ & 0.043 & 1.005\\
 & Random in $\pred$ & $-0.001$ & 0.079 & 1.005\\
 & Random bottom-50 & $-0.035$ & 0.174 & 1.045\\
\bottomrule
\end{tabular}
\caption{Norm-matched edits at $\alpha=0.5$ on the seed-17 anchor and branches: equal-target mean change in nats with a 5,000-resample bootstrap interval over the 30 target means for centroid rows, mean over eight draws with their standard deviation for random rows, and retention perplexity ratio.}
\label{tab:normmatched-full}
\end{table}

\section{Contrastive and matched-objective comparisons}\label{app:objectives}
\begingroup\sloppy
We compare two encoder backbones with their sentence-transformer variants, \nolinkurl{nreimers/MiniLM-L6-H384-uncased} with \nolinkurl{sentence-transformers/all-MiniLM-L6-v2}, and \nolinkurl{microsoft/mpnet-base} with \nolinkurl{sentence-transformers/all-mpnet-base-v2}, using their tied input embeddings in place of $W_U$, together with two decoders fine-tuned with InfoNCE, GTE-Qwen2-1.5B and E5-Mistral-7B, each compared with its cross-entropy base model (Table~\ref{tab:contrastive}). On the matched Mistral-7B-v0.1 base, SFT and InfoNCE both enrich the read pathway in $\inp$, with SFT higher in each of $W_Q$ and $W_K$ (Table~\ref{tab:mistral-objective}); the paired Wilcoxon tests run across the 32 layers of this single architecture.\par\endgroup

\begin{table}[h]
\centering
\small
\begin{tabular}{lcc}
\toprule
Checkpoint pair & Write/$\pred$ & Read/$\inp$\\
\midrule
MiniLM-L6: backbone to SBERT & 1.07 & 1.36\\
MPNet-base: backbone to SBERT & 1.07 & 1.92\\
GTE-Qwen2-1.5B: CE to InfoNCE & 1.10 & 1.28\\
E5-Mistral-7B: CE to InfoNCE & 1.71 & 1.28\\
Mistral-7B: SFT comparator & 1.13 & 1.77\\
\bottomrule
\end{tabular}
\caption{Contrastive fine-tuning deltas, backbone to sentence-transformer for the encoders and cross-entropy base to InfoNCE model for the decoders, with Mistral-7B SFT as the matched-architecture comparator, as normalised RSF.}
\label{tab:contrastive}
\end{table}

\begin{table}[h]
\centering
\small
\begin{tabular}{lcccc}
\toprule
Comparison & SFT & InfoNCE & Difference & Wilcoxon $p$\\
\midrule
$\inp$ ratio, mean of $W_Q,W_K$ & 1.769 & 1.282 & $+0.487$ & $<10^{-3}$\\
$\inp$ ratio, $W_K$ & 1.795 & 1.310 & $+0.485$ & $<10^{-3}$\\
$\inp$ ratio, $W_Q$ & 1.745 & 1.255 & $+0.490$ & $<10^{-3}$\\
\bottomrule
\end{tabular}
\caption{Read-pathway $\inp$ ratios on Mistral-7B-v0.1 under SFT and InfoNCE, with paired Wilcoxon tests across 32 layers.}
\label{tab:mistral-objective}
\end{table}

\section{Restricted adaptation}\label{app:lora}
LoRA adapters are trained on $W_Q,W_K$ (read-only), $W_O,W_2$ (write-only), $W_V,W_1$ (internal-only) or all six matrices, with the rank of each placement chosen so that the number of trainable parameters is the same across placements, since equal ranks give different parameter counts for matrices of different shapes. All placements share one SFT recipe and are run with two seeds; the loss gain is the reduction in held-out SFT loss, so that a positive value denotes lower loss, retention is the change in WikiText perplexity, and the RSF column of Table~\ref{tab:lora} measures the learned low-rank update against $\pred$.

\section{Downstream propagation of read and write edits}\label{app:propagation}
We inject a read-side or a write-side edit at $\ell^*=\lfloor2L/3\rfloor$, with the concept direction $u_{\mathcal C}$ shared between the two and the nominal strength $\alpha=0.5$. The write-side edit has the form of Equation~\ref{eq:edit}, and the read-side edit is $W\leftarrow W-\alpha(Wu_{\mathcal C})u_{\mathcal C}^\top$ for $W_Q$ and $W_K$. We measure
\begin{equation}
 M_1(\ell)=\|QK^\top_{\mathrm{edit}}(\ell)-QK^\top_{\mathrm{base}}(\ell)\|_F.
 \label{eq:propagation}
\end{equation}
At the edit layer, the write edit leaves the attention scores unchanged, since $W_O$ and $W_2$ act after attention has been computed, whereas the read edit changes them directly. In the layers after $\ell^*$, the ratio of read-edit to write-edit perturbation lies between $0.93$ and $0.97$ in every model and stage (Table~\ref{tab:propagation}), so in the downstream layers the two edits change the attention scores by similar amounts, whichever pathway they were applied to. A common unit direction and strength do not equalise the realised parameter norms or the activation perturbations, so the ratios compare the two edits at equal nominal strength.

\begin{table}[h]
\centering
\footnotesize
\setlength{\tabcolsep}{4pt}
\begin{tabular}{llrrcccc}
\toprule
Group & Stage & $L$ & $\ell^*$ & $M_1^W(\ell^*)$ & $M_1^R(\ell^*)$ & $M_1^W$ after & $M_1^R/M_1^W$ after\\
\midrule
Llama-3.2-3B & base & 28 & 18 & 0.0000 & 2.510 & 1.776 & 0.94\\
 & SFT & 28 & 18 & 0.0000 & 2.031 & 1.589 & 0.96\\
 & DPO & 28 & 18 & 0.0000 & 2.538 & 1.788 & 0.95\\
Llama-3-8B & base & 32 & 21 & 0.0000 & 1.792 & 2.011 & 0.94\\
 & SFT & 32 & 21 & 0.0000 & 1.754 & 1.965 & 0.95\\
 & DPO & 32 & 21 & 0.0000 & 1.701 & 1.907 & 0.94\\
 & ORPO & 32 & 21 & 0.0000 & 1.803 & 2.032 & 0.94\\
Tulu-3-8B & base & 32 & 21 & 0.0000 & 1.924 & 2.038 & 0.94\\
 & SFT & 32 & 21 & 0.0000 & 1.866 & 2.001 & 0.94\\
 & DPO & 32 & 21 & 0.0000 & 1.875 & 2.041 & 0.94\\
Qwen2.5-3B & base & 36 & 24 & 0.0000 & 2.155 & 3.674 & 0.97\\
 & SFT & 36 & 24 & 0.0000 & 2.137 & 3.770 & 0.96\\
 & ORPO & 36 & 24 & 0.0000 & 2.187 & 4.007 & 0.96\\
Qwen2.5-7B & base & 28 & 18 & 0.0000 & 2.936 & 96.95 & 0.93\\
 & SFT & 28 & 18 & 0.0000 & 2.962 & 96.45 & 0.94\\
 & DPO & 28 & 18 & 0.0000 & 2.973 & 96.40 & 0.94\\
Mistral-7B & base & 32 & 21 & 0.0000 & 4.372 & 2.305 & 0.97\\
 & SFT & 32 & 21 & 0.0000 & 4.972 & 2.504 & 0.96\\
 & DPO & 32 & 21 & 0.0000 & 2.332 & 1.895 & 0.97\\
 & ORPO & 32 & 21 & 0.0000 & 2.303 & 1.867 & 0.97\\
\bottomrule
\end{tabular}
\caption{Propagation of read and write edits. $M_1^W$ and $M_1^R$ are the values of Equation~\ref{eq:propagation} under write-only and read-only edits. The post-edit columns average over the layers following $\ell^*$.}
\label{tab:propagation}
\end{table}

\section{Activation-level comparison across architectures}\label{app:bridge}
For architectures without matched alignment deltas, we measure the raw activation fraction
\begin{equation}
 f_{\mathrm{act}}(h,\pred)=\frac{\|\pred h\|_2^2}{\|h\|_2^2}
 \label{eq:activation}
\end{equation}
at the median layer (Table~\ref{tab:act}). Autoregressive transformers place $0.149$--$0.163$ of the residual-stream energy in the prediction subspace, against $0.010$--$0.029$ for Dream-7B, where the fraction decreases monotonically with the evaluation noise level, and $0.010$ for Mamba-370M \citep{mamba}, while a randomly initialised transformer sits at $0.016$, close to its own $k/d$. Hidden dimension, normalisation, training data and architecture all enter this comparison, and the Dream sweep changes the evaluation distribution rather than the training history. Within autoregressive transformers, layer-level weight and activation concentration are positively correlated in four of five models (Table~\ref{tab:bridge}), with Qwen2.5-7B the exception.

\begin{table}[h]
\centering
\small
\begin{tabular}{llr}
\toprule
Family & Model/setting & Activation fraction\\
\midrule
Autoregressive transformer & Qwen2.5-7B & 0.163\\
 & Llama-3.1-8B & 0.157\\
 & Qwen2.5-0.5B & 0.156\\
 & Gemma-2-2B & 0.149\\
Dream-7B & $t=0.1$ & 0.029\\
 & $t=0.3$ & 0.028\\
 & $t=0.5$ & 0.025\\
 & $t=0.7$ & 0.019\\
 & $t=0.9$ & 0.010\\
State-space model & Mamba-370M & 0.010\\
Randomly initialised transformer & Llama-3.2-3B & 0.016\\
\bottomrule
\end{tabular}
\caption{Median-layer raw activation fraction in the rank-50 prediction subspace.}
\label{tab:act}
\end{table}

\begin{table}[h]
\centering
\small
\begin{tabular}{lrrr}
\toprule
Model & Layers & Spearman $\rho$ & $p$\\
\midrule
Llama-3.2-3B & 28 & 0.595 & $<10^{-3}$\\
Llama-3.1-8B & 32 & 0.625 & $<10^{-3}$\\
Qwen2.5-3B & 36 & 0.585 & $<10^{-3}$\\
Qwen2.5-7B & 28 & $-0.059$ & 0.77\\
Mistral-7B-v0.1 & 32 & 0.375 & 0.034\\
\bottomrule
\end{tabular}
\caption{Layer-level Spearman correlation between weight and activation RSF within autoregressive transformers.}
\label{tab:bridge}
\end{table}

\section{Attention-score gradient measurements}\label{app:selection-details}
\paragraph{Cohort and data}
The cohort contains Llama-3.2-1B, Llama-3.2-3B and Llama-3.2-3B-Instruct, OLMo-1B, Qwen2.5-3B and Qwen2.5-3B-Instruct, and Qwen3-0.6B, together with five public pretraining ladders, Amber-7B, OLMo-1B, Pythia-410M, Pythia-6.9B and Pythia-12B. Gradients are computed on short open-text sequences under teacher-forced next-token cross-entropy. Normalised gradient density and minibatch coherence use 200 sequences arranged as ten batches of twenty, and winner-sign statistics use 400 prompts from each of four corpora, WikiText-style English, HumanEval-style code, OPUS-Books French and GSM8K-style mathematical questions. Only positions that contribute to the loss are included, all runs except Pythia-12B (bfloat16) use fp32, and whole-prompt sign flips provide the permutation null for the sign statistics. The query--key group is $\{W_Q,W_K\}$ and the value--output group is $\{W_V,W_O\}$, with learned QK-normalisation parameters, normalisation gains and biases reported separately.

\paragraph{Statistics}
For each parameter group we store a per-parameter gradient statistic and normalise the group values within a run to sum to one, which gives the normalised gradient density $\nu$; it supports comparisons within a run and between checkpoints of the same architecture, while its absolute magnitude depends on the parameterisation. For minibatch gradients $\Gamma_b$ the minibatch coherence is
\begin{equation}
 \kappa=\frac{\|\overline\Gamma\|_F}{\mathbb E_b\big[\|\Gamma_b\|_F\big]},\qquad \overline\Gamma=\mathbb E_b[\Gamma_b],
 \label{eq:kappa}
\end{equation}
which measures directional agreement across batches. To resolve cancellation at the level of attention decisions, for every query position $i$ that contributes to the loss we take the winning key $j^\star$ and record $\operatorname{sign}([\partial\mathcal L/\partial s_i]_{j^\star})$. Within each head, the winner-sign coherence is the absolute signed imbalance divided by the number of decisions, averaged across heads, so a value of $0.05$ means that five percent of winner signs remain after cancellation; the global coherence pools all heads before taking the imbalance. Because the score gradient depends only on the scores, this statistic is unchanged by any reparameterisation of $W_Q$ and $W_K$ that leaves the scores unchanged.

\begin{table}[h]
\centering
\small
\begin{tabular}{lccccc}
\toprule
Model & $\nu_{\mathrm{QK}}$ & $\nu_{\mathrm{OV}}$ & $\kappa_{\mathrm{QK}}$ & $\kappa_{\mathrm{OV}}$ & Winner sign\\
\midrule
Llama-3.2-1B & $0.0161$ & $0.4128$ & $0.457$ & $0.586$ & $0.043$\\
Llama-3.2-3B & $0.0152$ & $0.3146$ & $0.435$ & $0.511$ & $0.043$\\
Llama-3.2-3B-Instruct & $0.0196$ & $0.3032$ & $0.594$ & $0.610$ & $0.057$\\
OLMo-1B & $0.0104$ & $0.8521$ & $0.524$ & $0.652$ & $0.041$\\
Qwen2.5-3B & $0.0209$ & $0.1362$ & $0.433$ & $0.519$ & $0.033$\\
Qwen2.5-3B-Instruct & $0.0274$ & $0.1472$ & $0.513$ & $0.571$ & $0.042$\\
\midrule
Qwen3-0.6B & $0.0058$ & $0.0120$ & $0.735$ & $0.712$ & $0.057$\\
\bottomrule
\end{tabular}
\caption{Static attention-score gradient statistics. $\nu$ is the normalised gradient density, $\kappa$ the minibatch coherence of Equation~\ref{eq:kappa}, and the winner-sign value the English per-head coherence. Qwen3 is shown separately because its learned QK normalisation holds $0.902$ of its normalised density budget.}
\label{tab:selection-static}
\end{table}

\paragraph{Cancellation within heads}
Table~\ref{tab:selection-static} lists the static values. Across all 28 model and corpus cells, per-head winner-sign coherence lies between $0.031$ and $0.059$, against permutation floors of $0.015$ to $0.024$, so between 94 and 97 percent of winner-score signs cancel within heads across a corpus, and the remaining imbalance lies above the permutation floor in almost every cell. Pooling heads before taking the imbalance gives a much smaller global coherence, at or below the floor in ten of the 28 cells, so the surviving consensus lives within heads rather than at the level of the whole model. Along the pretraining ladders of Table~\ref{tab:selection-ladders}, the per-head residual stays between 3.5 and 7.5 percent from the earliest to the latest stored checkpoint.

\begin{table}[h]
\centering
\small
\begin{tabular}{lcc}
\toprule
Ladder & Token span & Winner sign\\
\midrule
Amber-7B & 3.52B to 1.26T & $0.037\rightarrow0.059$\\
OLMo-1B & 4.0B to 3.09T & $0.035\rightarrow0.047$\\
Pythia-410M & 1.07B to 299.9B & $0.035\rightarrow0.056$\\
Pythia-6.9B & 1.07B to 299.9B & $0.039\rightarrow0.063$\\
Pythia-12B & 1.07B to 299.9B & $0.041\rightarrow0.065$\\
\bottomrule
\end{tabular}
\caption{Per-head winner-sign coherence along public pretraining ladders, first and last stored checkpoints. Pythia-12B was measured in bfloat16 and reaches $0.075$ before ending at $0.065$.}
\label{tab:selection-ladders}
\end{table}

\paragraph{Base and instruction-tuned checkpoints}
Instruction tuning raises the query--key density, from $0.0152$ to $0.0196$ in Llama-3.2-3B and from $0.0209$ to $0.0274$ in Qwen2.5-3B, and raises $\kappa_{\mathrm{QK}}$, from $0.435$ to $0.594$ and from $0.433$ to $0.513$. The value--output density changes from $0.3146$ to $0.3032$ and from $0.1362$ to $0.1472$, and $\kappa_{\mathrm{OV}}$ rises by less than $\kappa_{\mathrm{QK}}$ in both pairs, while value--output gradients remain denser and more coherent than query--key gradients, so the change happens within the ordering set by pretraining.

\paragraph{Learned QK normalisation}
In Qwen3-0.6B, learned QK normalisation absorbs $0.902$ of the normalised density budget, and the coherence ordering reverses, with $\kappa_{\mathrm{QK}}=0.735$ against $\kappa_{\mathrm{OV}}=0.712$. The derivative structure of Equation~\ref{eq:softmax} is unchanged, but the query--key statistics partly move into parameters that set the operating scale of the scores, which is why density and coherence are compared only between checkpoints with the same parameterisation.

\end{document}